%% file: main.tex
\documentclass[11pt, a4paper, copyright, nonumbering]{skywork}
\usepackage[authoryear, sort&compress, round]{natbib}
\usepackage{dblfloatfix}
\usepackage{ulem}
\usepackage{caption}
\usepackage{dramatist}
\usepackage{xspace}
\usepackage{pifont}
\usepackage{multirow}
\usepackage{adjustbox}
\usepackage{tcolorbox}
\usepackage{xltabular}
\usepackage{longtable}
\usepackage{hyperref}
\usepackage{makecell}
\usepackage{algorithm}
\usepackage[noend]{algpseudocode}
\usepackage{amsfonts}
\usepackage{amsmath}
\usepackage{amssymb}
\usepackage{lineno}
\usepackage[bottom]{footmisc}
\usepackage{CJKutf8}
\usepackage{subfigure}
\usepackage{setspace}
\usepackage{titlesec}
\usepackage{enumitem}
\usepackage[framemethod=TikZ]{mdframed}
\usepackage{lipsum}
\usepackage{amsmath}
\usepackage{colortbl} 
\usepackage{booktabs} 
\usepackage[english]{babel}
\usepackage{caption}
\usepackage{subcaption}
\usepackage{epigraph}
\usepackage{wrapfig}
\usepackage{tablefootnote}
\usepackage{float}
\usepackage{dashrule}
\usepackage{arydshln}

\definecolor{greyrow}{rgb}{0.9, 0.9, 0.9} 
\definecolor{orangerow}{rgb}{1, 0.95, 0.9}
\definecolor{greenrow}{rgb}{0.9, 1, 0.9}
\definecolor{bluerow}{rgb}{0.9, 0.9, 1}
\definecolor{redrow}{rgb}{1, 0.9, 0.9}

\makeatletter
\def\@BTrule[#1]{%
  \ifx\longtable\undefined
    \let\@BTswitch\@BTnormal
  \else\ifx\hline\LT@hline
    \nobreak
    \let\@BTswitch\@BLTrule
  \else
     \let\@BTswitch\@BTnormal
  \fi\fi
  \global\@thisrulewidth=#1\relax
  \ifnum\@thisruleclass=\tw@\vskip\@aboverulesep\else
  \ifnum\@lastruleclass=\z@\vskip\@aboverulesep\else
  \ifnum\@lastruleclass=\@ne\vskip\doublerulesep\fi\fi\fi
  \@BTswitch}
\makeatother

\addto\extrasenglish{
}

 {\begin{list}{}%
         {\setlength{\leftmargin}{#1}}%
         \item[]%
 }
 {\end{list}}
 
\reportnumber{001} 

\renewcommand{\today}{}

\vspace{-5mm}
\title{\centering Skywork-Math: Data Scaling Laws for Mathematical Reasoning in Large Language Models --- The Story Goes On}
\vspace{-5mm}
\author[*]{
\normalsize
\hspace{5em}
\textbf{Liang Zeng, Liangjun Zhong, Liang Zhao, Tianwen Wei}, \newline Liu Yang, Jujie He, Cheng Cheng, Rui Hu, Yang Liu, Shuicheng Yan, Han Fang, Yahui Zhou 

\normalsize \{forename\}.\{surname\}@kunlun-inc.com \;\;\;\;\;\;
Skywork AI, Kunlun Inc.
}
\correspondingauthor{}

\input{commands.tex}

\input{sections/abstract}

\begin{document}
\begin{CJK*}{UTF8}{gbsn}

\maketitle

\input{sections/introduction}
\input{sections/related}
\input{sections/method}
\input{sections/experiment}
\newpage
\input{sections/conclusion}
\input{sections/acknowledgement}
\newpage
\bibliography{main}

\newpage
\appendix
\input{sections/appendix}

\end{CJK*}
\end{document}

%% file: commands.tex
\renewcommand{\phi}{\varphi}

\renewcommand{\epsilon}{\varepsilon}
\renewcommand{\imath}{\mathrm{i}}

\newlength{\restsubwidth}
\newlength{\restsubheight}
\newlength{\restsubmoreheight}
\newcommand{\rest}[2]{%
        \settowidth{\restsubwidth}{\ensuremath{#2}}
        \settoheight{\restsubheight}{\ensuremath{{}_{#2}}}
        \ensuremath{{#1\hskip 0.5pt}_{\vrule\kern2pt\parbox[b][%
        4pt][b]{\the\restsubwidth}{%
                        \ensuremath{{}_{#2}}}}}
        }

%% file: sections/abstract.tex
\begin{abstract}
In this paper, we investigate the underlying factors that potentially enhance the mathematical reasoning capabilities of large language models~(LLMs). We argue that the data scaling law for math reasoning capabilities in modern LLMs is far from being saturated, highlighting how the model's quality improves with increases in data quantity. To support this claim, we introduce the Skywork-Math model series, supervised fine-tuned (SFT) on common 7B LLMs using our proposed 2.5M-instance Skywork-MathQA dataset. Skywork-Math 7B has achieved impressive accuracies of 51.2\% on the competition-level MATH benchmark and 83.9\% on the GSM8K benchmark using only SFT data, outperforming an early version of GPT-4 on MATH. The superior performance of Skywork-Math models contributes to our novel two-stage data synthesis and model SFT pipelines, which include three different augmentation methods and a diverse seed problem set, ensuring both the quantity and quality of Skywork-MathQA dataset across varying difficulty levels. Most importantly, we provide several practical takeaways to enhance math reasoning abilities in LLMs for both research and industry applications.
\end{abstract}

%% file: sections/introduction.tex
\begin{figure}[htbp]
    \centering    \includegraphics[width=0.57\textwidth]{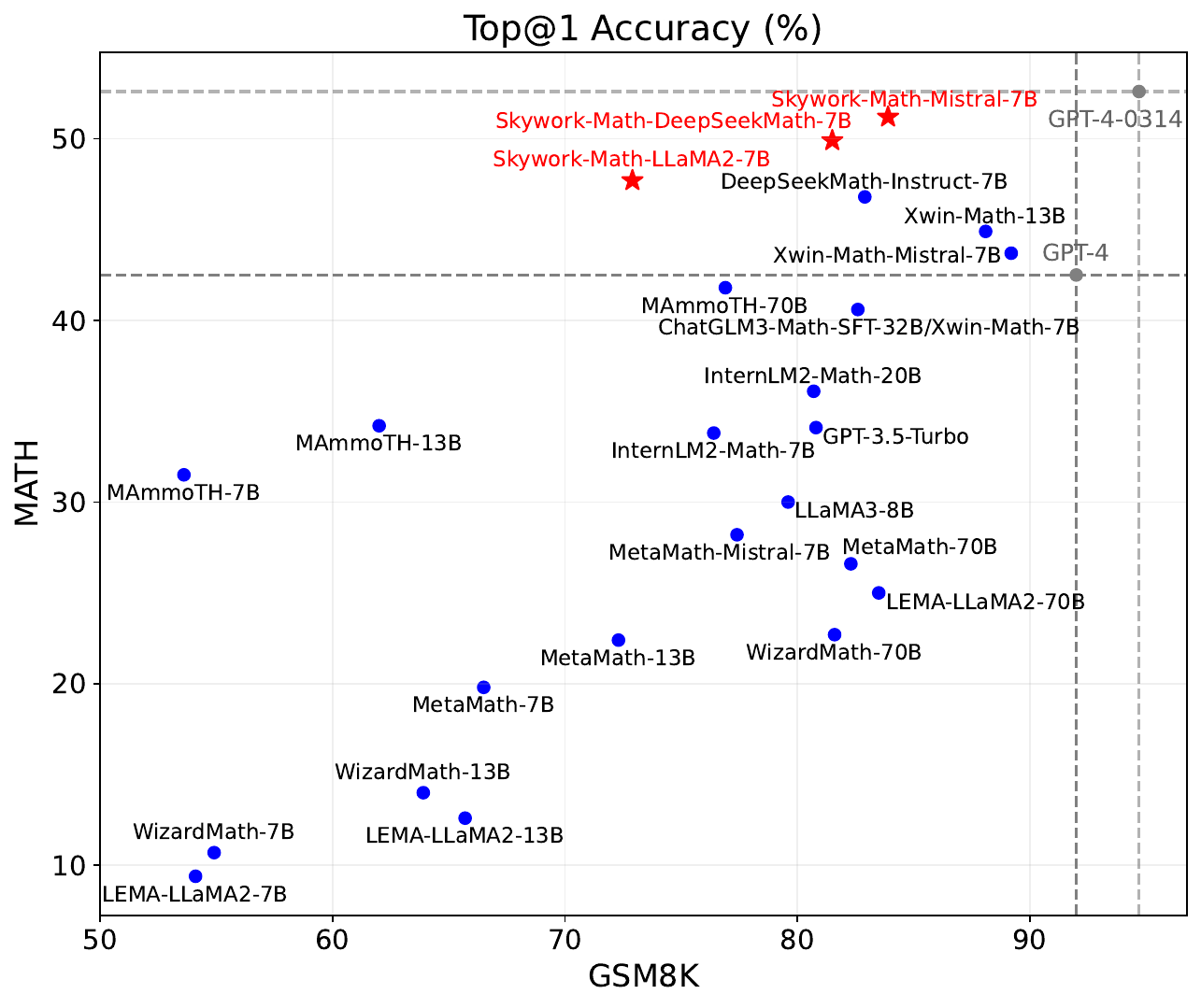}
    \caption{Top1 accuracy on GSM8K~\citep{cobbe2021trainingGSM8K} and MATH~\citep{hendrycks2021measuring} using only SFT techniques, without using external toolkits and voting techniques. Following MetaMath~\citep{yu2023metamath}, we employ a zero-shot chain-of-thought evaluation framework. Skywork-Math models achieve state-of-the-art accuracy among models smaller than 10B parameters using only synthetic SFT data and surpass an early version of GPT-4 on MATH.}
    \label{acc_all_models}
\end{figure}

\section{Introduction}

\epigraph{\textit{More is different.}}{----Philip W. Anderson, 1972}

Reasoning ability is a hallmark of human intelligence~\citep{huang2022towards, gendron2024LLMnotReasoner, wei2023chainofthought}. Although Large Language Models~(LLMs) have recently demonstrated significant capabilities in various tasks such as conversation~\citep{openai2024gpt4, claude3modelcard, gpt3.5} and summarization~\citep{wei2023skywork,  yang2023baichuan2, workshop2023bloom, almazrouei2023falcon}, they often struggle with complex reasoning tasks~\citep{gendron2024LLMnotReasoner, lu2023chameleon, wu2024reasoningorreciting}. One particularly challenging area is mathematical reasoning~\citep{hendrycks2021measuring, cobbe2021trainingGSM8K, zhong2023agieval, arora2023llmsJEEBench, he2024olympiadbench}, which requires the ability to solve mathematical problems and derive logical conclusions in a step by step manner~\citep{wei2023chainofthought, saxton2019analysingDeepmindMath, shao2024deepseekmath, yu2023metamath, toshniwal2024openmathinstruct1}.

Two prevailing beliefs guide researchers and practitioners in enhancing mathematical reasoning abilities of LLMs. The first belief posits that complex reasoning abilities, especially mathematical reasoning, are emergent abilities that exist in large language models but not in small models~\citep{wei2023chainofthought, wei2022emergent}. Typically, models with more than 30 billion parameters exhibit the strong mathematical reasoning ability~\citep{brown2020language}. The second belief is the seminal "superficial alignment" hypothesis~\citep{zhou2023lima}, which asserts that \textit{"A model’s knowledge and capabilities are learnt almost entirely during pre-training, while alignment teaches it which sub-distribution of formats should be used when interacting with users."}. According to this hypothesis, the alignment process, primarily through supervised fine-tuning~(SFT), does not inject new knowledge or improve inherent abilities but rather adjusts the output response format. This implies that the strong mathematical reasoning ability may not be significantly improved by a large amount of synthetic SFT data. 

In this paper, we re-examine these two common beliefs mentioned above regarding mathematical reasoning abilities of LLMs. For the first belief, we introduce the Skywork-Math model series, which are supervised fine-tuned~(SFT) on common 7B pre-trained LLM models without employing other complex alignment techniques such as RLHF~\citep{bai2022training,casper2023open} and DPO~\citep{rafailov2024direct}. Skywork-Math 7B models have achieved impressive accuracies of 51.2\% on the competition-level MATH~\citep{hendrycks2021measuring} benchmark and 83.9\% on the GSM8K~\citep{cobbe2021trainingGSM8K} benchmark, notably outperforming an early version of GPT-4 on MATH. Our empirical findings, consistent with the conclusions in~\cite{li2024common}, suggest that strong mathematical reasoning ability can indeed exist in common 7B language models. Moreover, scaling up synthetic SFT data can further enhance the mathematical reasoning ability of Skywork-Math 7B models.

For the second belief, we propose Skywork-MathQA high-quality SFT dataset containing 2.5 million instances, which is much larger than open-sourced dataset of its kind to date, such as MetaMathQA~\citep{yu2023metamath} containing 395K samples.
We empirically observe that the scaling law curve on the SFT alignment for mathematical reasoning in modern LLMs is far from being saturated~(ref. Figure~\ref{scale_figure}).
We have carefully scaled the Skywork-MathQA SFT dataset with diverse and high-quality samples specifically within the mathematical domain to enhance the model’s capability in understanding and solving mathematical problems.

Due to the scarcity of high-quality and challenging mathematical data, various pipelines and prompts have been employed to generate synthetic mathematical data~\citep{yu2023metamath, shao2024deepseekmath, li2024common, toshniwal2024openmathinstruct1, wei2023chainofthought,wang2023selfconsistency}. To address this deficiency, we employ GPT-4 to generate a substantial amount of synthetic data through a novel two-stage data synthesis pipeline, in conjunction with the corresponding model SFT process. 
In stage 1, our objective is to obtain normal synthetic problems to enhance the models' general comprehension of mathematical problems. To maintain the diversity in data selection process, we utilize the core-set approach~\citep{sener2018activeCoreSet} on enlarged seed problems. However, as the data volume increases, we empirically observe that the relationship between performance and data quantity begins to plateau. Accordingly, in stage 2, we diversify the dataset further by introducing a proportion of augmented hard problems~(ref. Figure~\ref{data_exp} for illustrative examples), thereby exposing the model to more challenging mathematical questions.
Without continual pre-training on a large-scale math corpus~\citep{shao2024deepseekmath,azerbayev2024llemma}, Skywork-Math models achieve impressive performance with just supervised fine-tuning on common pre-trained LLMs containing only 7B parameters. 

Most importantly, we provide valuable insights and practical takeaways to enhance the mathematical reasoning ability in LLMs, benefiting both research and industry communities:

\begin{tcolorbox}[colback=blue!3!white, colframe=blue!75!black, title=Highlighted Takeaways]
\begin{itemize}[left=0pt]
    \item The potential for math reasoning capabilities in modern LLMs is far from exhausted. The quality of LLMs can significantly improve with increases in data quantity~(ref. Figure~\ref{scale_figure}). Skywork-Math 7B models already demonstrate strong mathematical reasoning abilities by SFTing on common 7B pre-trained LLM models.
    \item The learning process for accessing the math reasoning ability involves multiple stages. Training LLM models in a meaningful order, from the easy problems to the hard ones, can provide performance improvements.
    \item When scaling the synthetic SFT dataset, increasing the diversity of seed problems and augmentation methods can improve the math reasoning performance of LLMs.
    \item Selecting influential data with high-quality from a large dataset is non-trivial~\citep{engstrom2024dsdm}. Our empirical results indicate that some straightforward methods to select the so-called "high-quality" data may not increase~(and can even hurt) LLMs' performance compared to randomly selecting data. The selection process involves multiple constraints, and the "high-quality" data could significantly decrease the difficulty level of problems, thus negatively impacting the performance of LLMs.
    \item The LLM models have strong knowledge transfer capabilities for mathematical reasoning across bilingual benchmarks~(i.e., English and Chinese). We hypothesize that this can be attributed to the inherent nature of symbols and numbers in math problems, which retain their intrinsic meaning regardless of the language used.
    \item Although Skywork-Math 7B models have achieved considerable improvement in robustness tests compared to other open-source LLM models, they remain sensitive to the distractors in math word problems compared with proprietary GPT-4 models.
    \item Sparse MOE models cannot clearly exceed the performance upper bound of their dense counterparts through SFT alignment in the context of math reasoning.
    \item Two subtle but crucial practical implementation techniques---preventing data leakage and considering the influence of model maximum length---significantly impact the final performance of LLM models.
\end{itemize}
\end{tcolorbox}

%% file: sections/related.tex
\section{Related Work}
\paragraph{Alignment in LLMs.}
Large Language Models~(LLMs) have recently transformed Natural Language Processing~(NLP)~\citep{openai2024gpt4, anil2023palm2, claude3modelcard,touvron2023llama}, excelling in tasks such as automated summarization~\citep{workshop2023bloom} and machine translation~\citep{almazrouei2023falcon}.
Alignment in LLMs refers to the process of ensuring that the model's outputs adhere to user preferences~\citep{shen2023large}. Various techniques contribute to achieving alignment, including supervised fine-tuning~(SFT)~\citep{alpaca}, reinforcement learning from human feedback~(RLHF)~\citep{bai2022training}, and direct policy optimization~(DPO)~\citep{rafailov2024direct}. Among these techniques, SFT is typically an indispensable method for aligning LLMs and has achieved highly competitive performance across various tasks~\citep{vicuna2023}, particularly in mathematical reasoning~\citep{li2024common}. SFT involves fine-tuning a pre-trained large model using annotated data, making the model's performance more accurate for downstream tasks. Our work aims to deeply explore the performance boundaries of common 7B pre-trained LLMs using only the SFT alignment technique.

\paragraph{Quantity and Quality of SFT Data.}
Data is the fuel that powers the performance of LLMs. This ongoing discussion about whether the quantity or quality of SFT data is more important highlights their significance in enhancing the SFT performance of LLMs. 
(1) \textbf{Quantity.} Many recent research demonstrates the scaling properties in LLM fine-tuning~\citep{kaplan2020scalingLaws,li2024common}. The size of the fine-tuning dataset is a crucial factor affecting the LLMs' performance. However, the optimal fine-tuning data size is highly task-dependent~\citep{zhang2024scalingFinetuning}. 
(2) \textbf{Quality.} Several studies~\citep{li2024quantitytoQuality, cao2023instructionMining, zhou2023lima, gunasekar2023textbooks} argue that the quality of fine-tuning data is equally critical. The renowned "less is more" work~\citep{zhou2023lima} suggests that substantial knowledge acquisition occurs during the pre-training stage, minimizing the need for extensive fine-tuning data. Additionally, the Instruction-Following Difficulty~(IFD) metric introduced by~\citep{li2024quantitytoQuality} and the QaDS strategy proposed in~\citep{ni2024exploring} aim to select diverse and high-quality instruction-following data to enhance LLM fine-tuning efficiency. Collecting a huge number of high-quality mathematical reasoning data is often time-consuming and labor-intensive. In this work, we generate a substantial amount of SFT synthetic data to investigate how the quantity of data impacts the performance of LLM models in mathematical reasoning.

\paragraph{Mathematical Reasoning in LLMs.}
LLMs have recently achieved significant progress in the area of mathematical reasoning~\citep{shao2024deepseekmath}.
Initial benchmarks, such as simple math problems~\citep{saxton2019analysingDeepmindMath,lan2022mwptoolkit}, were readily solved by recent LLM models. This success prompts the introduction of more challenging benchmarks, such as GSM8K~\citep{cobbe2021trainingGSM8K} and MATH~\citep{hendrycks2021measuring}. 
Many recent works have proposed continual pre-training on massive math corpora to improve their math reasoning capabilities~\citep{paster2023openwebmath, azerbayev2024llemma, shao2024deepseekmath, jiang2023mistral}.
Furthermore, significant progress has been made in alignment for solving mathematical problems~\citep{ni2024exploring, yue2023mammoth, xu2024chatglmmath, shao2024deepseekmath, yu2023metamath, luo2023wizardmath, li2024common}. These studies focus on generating high-quality synthetic data or collecting human-labeled data for model fine-tuning and alignment in the domain of math problem-solving.
Additionally, reasoning frameworks aim at improving math capacity, such as the chain-of-thought~(COT) prompting technique~\citep{wei2023chainofthought,wang2023selfconsistency}, which enable LLMs to break down the reasoning process into manageable steps, resulting in more accurate outputs. 
Moreover, some complex math problems need the ability to conduct accurate arithmetic operations, a capability that LLMs often lack~\citep{yuan2023largeArithmetic}. For tool-integrated math problem-solving, program-of-thoughts~\citep{chen2023programofthoughts, shao2024deepseekmath, toshniwal2024openmathinstruct1} prompts LLMs to produce answers in the code format, which are then executed by a code interpreter.
Preliminary work indicates that SFT can improve the performance of open-source LLMs on mathematical reasoning tasks by fine-tuning them on synthetic data~\citep{yu2023metamath,li2024common}. Building on this foundation, our work aims to thoroughly investigate the performance limits of common 7B pre-trained LLMs using only SFT synthetic data. We seek to determine the extent to which data quantity impacts LLM quality and to understand the mechanisms behind this influence.

%% file: sections/method.tex
\begin{figure}[t]
    \centering
    \includegraphics[width=0.8\textwidth]{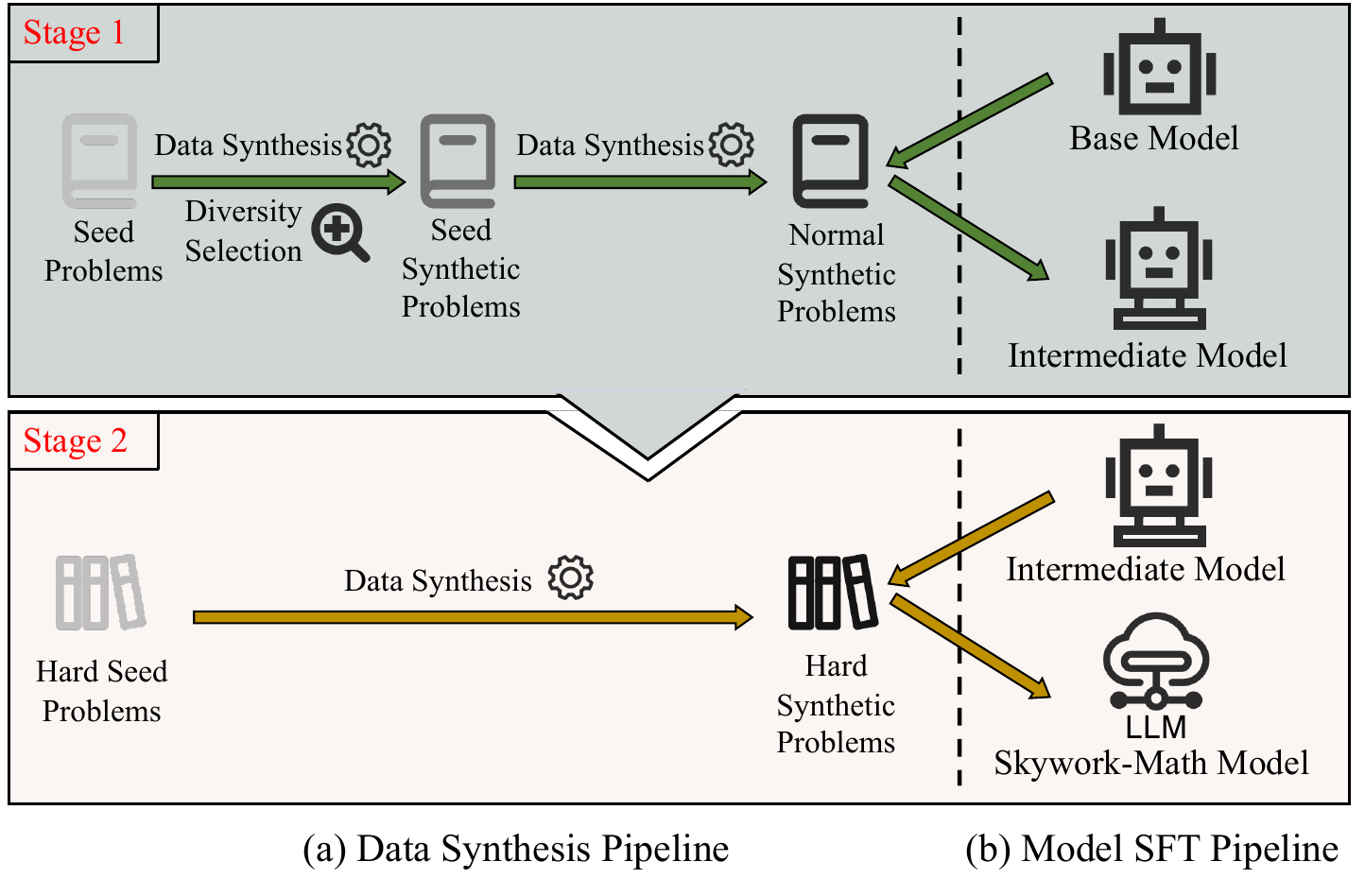}
    \caption{Overview of our proposed two-stage method. (a) The data synthesis pipeline of the Skywork-MathQA dataset. (b) The model SFT pipeline of the Skywork-Math model series.}
    \label{fig:framework}
\end{figure}

\section{Method}
In this section, we present the detailed methodology of Skywork-Math 7B models, as illustrated in Figure~\ref{fig:framework}. Skywork-Math models aim to enhance math reasoning abilities during the model alignment process, particularly in the SFT stage, using common and publicly available 7B pre-trained models. We employ a two-stage SFT approach, in conjunction with two data synthesis pipelines to produce high-quality data. 
In stage 1, we feed base pre-trained models with our generated normal synthetic problems to produce an intermediate model. In stage 2, to mitigate the diminishing returns in LLMs' performance as the quantity of data increases, we generate hard synthetic problems and develop our Skywork-Math models.
To ensure the quality of data, we primarily utilize GPT-4~\footnote{Without further clarification, the version of GPT-4 used in this paper is GPT-4-1106-preview.}~\citep{openai2024gpt4} to generate 2.5M-instance synthetic Skywork-MathQA dataset.

\paragraph{Supervised Fine-Tuning~(SFT).}
SFT is an important and widely-used alignment technique in LLMs to enhance pre-trained models for excelling at specific tasks~\citep{shen2023large}. 
We denote the token space of an input query and output response as $\mathcal{X}$ and $\mathcal{Y}$, respectively. Typically, LLMs generate an output response sequence $\mathbf{y} = (y_1, y_2, \ldots, y_T)$ in response to a given prompt query $\mathbf{x} = (x_1, x_2, \ldots, x_n)$. LLMs are the auto-regressive models characterized by a conditional probability distribution parameterized by $\theta$ as
\begin{equation}
    \mathbb{P}_\theta(\mathbf{y} \mid \mathbf{x}) = \prod_{t=1}^{T} \mathbb{P}_\theta(y_t \mid \mathbf{x}, y_{1:t-1}).
\end{equation}
Let a mathematical reasoning SFT training dataset be $\mathcal{D} = \{(\mathbf{x}^i, \mathbf{y}^i)\}_{i=1}^N$, where $\mathbf{x}^i$ and $\mathbf{y}^i$ represent the $i$-th query and response, respectively~\footnote{In what follows, we use the terms query-response and question-answer pairs interchangeably.}. Here, $N$ is the total quantity of the SFT training dataset. Given such a dataset $\mathcal{D}$, SFT can be performed using the following cross-entropy loss:
\begin{equation}
    \mathcal{L}(\theta) = -\frac{1}{N} \sum_{i=1}^{N} \sum_{t=1}^{T} \log \mathbb{P}_\theta (y_t^i \mid \mathbf{x}^i, \mathbf{y}_{1:t-1}^i).
\end{equation}

\paragraph{Seed Problems.} We adopt publicly available high-quality mathematical datasets to generate our Skywork-MathQA dataset. To prevent data leakage in the testing phase, we only use the training sets from data sources. The data sources are as follows:
\begin{itemize}
    \item \textbf{MATH}~\citep{hendrycks2021measuring} contains high school-level mathematical problems, some of which are from competitions such as the AIME and AMC. This dataset consists of 7,500 training data entries. Solving these problems requires advanced reasoning abilities and a comprehensive mathematical knowledge base. This dataset categorizes problems into five levels of difficulty and seven subdomains of high school mathematics. 
    \item We also use \textbf{other data sources} as seed problems. These included non-proving problems from OlympiadBench~\citep{he2024olympiadbench}, mathematical problems from AGIEval~\citep{zhong2023agieval} benchmark, and various problems in calculus, differential, statistics domains from SciBench~\citep{wang2024scibench} and JEEBench~\citep{arora2023llmsJEEBench}.
\end{itemize}
Here we do not use the training set of GSM8K as the seed problems because: (1) Math word problems represent a narrow category compared to general math problems~\footnote{Typically, math word problems involve a mathematical exercise presented in a narrative form, requiring the extraction of numbers from the text and performing a sequence of elementary calculations using basic arithmetic operations~$(+-×÷)$ to reach the final answer.}, and an excessive focus on math word problems may reduce the diversity of the synthetic SFT data. (2) We empirically find that the requirements of math reasoning ability to solve the easy problem in the MATH benchmark are relatively equivalent to those needed for GSM8K.

\begin{figure}[htbp]
\begin{tcolorbox}[colback=blue!3!white, colframe=RoyalPurple!55!Aquamarine!100!, title=Normal Problem]
\textbf{Query}: What is the domain of the function $f(x) = \frac{3x+1}{x+8}$? Enter your answer using interval notation.

\hdashrule[0.5ex]{\linewidth}{1pt}{3pt}

\textbf{Response}: 
The function $f(x) = \frac{3x+1}{x+8}$ is defined for all real numbers except where the denominator is zero. Since the denominator $x+8$ is zero when $x=-8$, we exclude this value from the domain.
Thus, the domain of $f(x)$ in interval notation is $(-\infty, -8) \cup (-8, \infty)$.

The answer is $(-\infty, -8) \cup (-8, \infty)$.
\end{tcolorbox}

\begin{tcolorbox}[colback=blue!3!white, colframe=RoyalPurple!55!Aquamarine!100!, title=Hard Problem]
\textbf{Query}: Let $f(x) : \mathbb{R} \to \mathbb{R}$ be a function such that
\[\frac{f(x) f(y) - f(xy)}{3} = x + y + 2\]
for all $x,$ $y \in \mathbb{R}.$  Find $f(x).$

\hdashrule[0.5ex]{\linewidth}{1pt}{3pt}

\textbf{Response}: 
Setting $y=1$ gives us
\[\frac{f(x)f(1) - f(x)}{3} = x + 1 + 2,\]
which simplifies to
\[f(x)(f(1) - 1) = 3x + 9.\]
Since $f(x)$ is a function that maps $\mathbb{R}$ to $\mathbb{R}$, and since this holds for all $x \in \mathbb{R}$, $f(1) - 1$ cannot be zero because otherwise the right-hand side of the equation would always be $0$, which is not true. Thus, we can divide both sides of the equation by $f(1) - 1$ to get $$f(x) = \dfrac{3x + 9}{f(1) - 1}.$$
Setting $x=1$ in this equation gives us
$f(1) = \dfrac{3(1) + 9}{f(1) - 1},$
which is equivalent to $f(1)(f(1) - 1) = 12.$
Let's denote $f(1)$ as $a$, so we have
$a(a - 1) = 12.$
This is a quadratic equation in $a$: $a^2 - a - 12 = 0.$
The solutions to this quadratic equation are $a = -3$ or $a = 4$. Since $a$ represents $f(1)$ and we have established that $f(1) - 1$ cannot be zero, we discard the solution $a = 1$ and accept $a = 4$. Therefore, $f(1) = 4$ and the function $f(x)$ is
\[f(x) = \frac{3x + 9}{f(1) - 1} = \frac{3x + 9}{4 - 1} = \frac{3x + 9}{3} = x + 3.\]
The answer is x + 3.

\end{tcolorbox}
\caption{Two examples of query-response pairs in the Skywork-MathQA dataset. The top figure illustrates a normal problem, and the bottom figure depicts a hard problem. }
\label{data_exp}
\end{figure}

\paragraph{Synthesis Process.}
\label{sec:synthesis}
We aim to answer the following question: as we gradually increase the quantity $N$ of the Skywork-MathQA dataset, does the models' math reasoning ability improve correspondingly? For a given query/problem $\mathbf{x}^i$, particularly the challenging competition-level math problems, manually annotating the response/answer $\mathbf{y}^i$ is time-consuming and often infeasible for non-experts due to the required specific domain knowledge. 
Therefore, we utilize the top-performing GPT-4 models to synthesize diverse, high-quality SFT data~\citep{li2024common}. The data synthesis process in the Skywork-MathQA dataset consists of two stages. In stage 1, we generate 2.1 million normal synthetic problems. In stage 2, we further generate 0.4 million hard synthetic problems, increasing the Skywork-MathQA dataset to a total of 2.5 million instances.
Note that all data samples in the Skywork-MathQA dataset strictly adhere to the same data format. We instruct the Skywork-Math models to use the prefix "\textbackslash nThe answer is " before generating answers in their responses. Figure~\ref{data_exp} presents two examples from our Skywork-MathQA dataset: one is a normal problem, and the other is a hard problem. In the following sections, we will introduce the two-stage data synthesis pipeline along with its model SFT process.

\subsection{Stage 1: Normal Synthetic Problems}
In this stage, we examine how the quality of Skywork-Math models improves as the quantity of SFT data increases. We generate 2.1 million high-quality and diverse SFT data within math reasoning domains by GPT-4. Our primary goal is to equip the model with a comprehensive understanding of mathematical reasoning problems by exposing it to a diverse range of math questions. Our empirical findings indicate that diversity is crucial for generating and scaling SFT data~(ref. Section~\ref{exp:diversity}). We investigate this issue from two perspectives: data augmentation methods and diversity selection of seed problems.

\paragraph{Data Augmentation Methods.}
\label{sec:data_augmentation}
To ensure diversity in our synthetic data, we employ three distinct methods to augment our Skywork-MathQA dataset. We notice that the differences among these augmentation methods are subtle, however, combining these methods to improve diversity indeed influences the model's performance. 
Three data augmentation methods have distinct approaches. By combining them, we can leverage the advantages of all three unique approaches in our data synthesis pipeline. Figure~\ref{prompt:snippet} demonstrates three prompt snippets used in our paper to highlight the characteristics of these distinct approaches.
Detailed examples of the same query with different responses using these three methods can be found in Appendix~\ref{appendix:aug_exp}.

The first data augmentation method we adopt is MetaMathQA~\citep{yu2023metamath}, which comprises four specific approaches: three for query bootstrapping and one for response augmentation. For \textit{query augmentation}, we leave the corresponding query unchanged and employ GPT-4 to refine its response. For query bootstrapping, the \textit{rephrasing} method utilizes pre-defined prompts to generate more questions, followed by the few-shot Chain-of-Thought~(COT)~\citep{wei2023chainofthought} prompting to generate answers. Additionally, the \textit{FOBAR}~\citep{jiang2024forwardbackwardFOBAR} and \textit{self-verification}~\citep{weng2023largeSelfVerification} methods deterministically convert the problem into a backward format to mimic backward reasoning, i.e., given the result and think backward to determine the unknown variable in the question. After transforming the questions, we then generate corresponding answers with COT techniques using GPT-4. We also strive to balance the quantity of SFT data produced by these four augmentation approaches.

The second data augmentation method is the \textit{Evol-Instruct} approach, as implemented in WizardLM~\citep{xu2023wizardlm}. Starting from the initial set of mathematical problems, \textit{Evol-Instruct} iteratively rewrites them step by step into more complex queries. We set the maximum length of the evolutionary trajectory to five steps and employ the following five augmentation strategies:
\begin{itemize}[]
    \item Rewrite the original problem to create a completely new problem of similar length and difficulty.
    \item Add constraints and requirements to the original problem.
    \item Increase the complexity of the original problem in both depth and breadth.
    \item Replace general concepts with more specific ones.
    \item Explicitly request additional steps in the reasoning process of the original question.
\end{itemize}

\begin{figure}[H]
\begin{tcolorbox}[colback=blue!3!white, colframe=RoyalPurple!55!Aquamarine!100!, title=Prompts for MetaMath with Backward Reasoning, fontupper=\small]
You are an AI assistant to help me solve math problem. \textcolor{red}{I first give you an original question with an unknown variable $X$, then I give you the answer to this original question. You help me determine the value of $X$.} Follow the given examples and answer the question.

\textbf{Query:} On a map, a 12-centimeter length represents $X$ kilometers. How many kilometers does a 17-centimeter length represent? \#\#\# If we know the answer to the above question is 102, what is the value of unknown variable $X$?

\textbf{Response:} We know that on the map, a 12-centimeter length represents $X$ kilometers. To find out how many kilometers a 17-centimeter length represents, we can set up a proportion using the given information: 12 centimeters / $X$ kilometers = 17 centimeters / 102 kilometers. We can cross-multiply and solve for $X$: 12 * 102 = 17 * $X$, 1224 = 17 * $X$. Dividing both sides by 17, we get: $X$ = 72
The value of X is 72.

\textbf{.....}  \textcolor{blue}{[Few-shot examples]}
\end{tcolorbox}

\begin{tcolorbox}[colback=blue!3!white, colframe=RoyalPurple!55!Aquamarine!100!, title=Prompts for Evol with Evol-Instruct, fontupper=\small]
I want you act as a Prompt Rewriter. \textcolor{red}{Your objective is to rewrite a given prompt into a more complex version to make those famous AI systems (e.g., ChatGPT and GPT4) a bit harder to handle.} But the rewritten prompt must be reasonable and must be understood and responded by humans. \textcolor{red}{Please add one more constraints/requirements into \#Given Prompt\#.}

\textbf{......}  \textcolor{blue}{[Omit some specific rules]}

\#Given Prompt\#:
<Here is instruction.>

\#Rewritten Prompt\#:
\end{tcolorbox}

\begin{tcolorbox}[colback=blue!3!white, colframe=RoyalPurple!55!Aquamarine!100!, title=Prompts for Xwin with Self-Correction, fontupper=\small]
Please act as a professional math teacher. Your goal is to create high quality math word problems to help students learn math. You will be given a math question. Please create a new question based on the Given Question and following instructions. To achieve the goal, you have three jobs.

\textbf{......}  \textcolor{blue}{[Omit some specific rules]}

\textcolor{red}{VERIFICATION AND MODIFICATION: <solve the question step-by-step and modify it to follow all principles>}

FINAL CREATED QUESTION: <your final created question>

\end{tcolorbox}
\caption{Prompt snippets for MetaMath~\cite{yu2023metamath}, Evol~\cite{luo2023wizardmath}, and Xwin~\cite{li2024common} are showcased, with their distinct approaches highlighted in red. The prompts are mainly derived from the original papers with minor modifications. For the sake of brevity, some specific few-shot examples and rules have been omitted.}
\label{prompt:snippet}
\end{figure}

The third data augmentation method is \textit{question generation with self-correction}, as practiced in Xwin \citep{li2024common}. Specifically, we instruct GPT-4 to refine the input question and then verify it step-by-step to assess its logical and mathematical consistency. If the question is found to be imperfect, we instruct the GPT-4 to modify it based on the verification results.

\paragraph{Diversity Selection of Seed Problems.}
\label{sec:diversity_selection}
Initially, we simply use the training dataset of MATH along with additional mathematical data from other sources as the seed problem to generate queries and responses. 
To improve the diversity of seed problems, we employ the core-set approach~\citep{sener2018activeCoreSet}, which selects a representative subset of data that maximizes diversity while maintaining coverage of the original dataset's key features.
As shown in Figure~\ref{fig:framework}, we first perform data synthesis on the initial seed problems and then apply the core-set approach~\citep{sener2018activeCoreSet} to obtain seed synthetic problems. We further perform data synthesis on these seed synthesis problems to get the normal synthetic problems with 2.1 million instances.
We select common 7B pre-trained LLMs as base models and fine-tune these models on normal synthetic problems to produce the intermediate models with a general understanding of various mathematical problems and concepts.

\subsection{Stage 2: Hard Synthetic Problems}
As the quantity of data increased, we empirically observe that the relationship between performance and data quantity begins to plateau~(ref. Section~\ref{sec:level}). Motivated by the concept of curriculum learning~\citep{soviany2022curriculum, bengio2009curriculum}, we recognize that models can learn much better when data are organized in a meaningful order rather than presented randomly, introducing more complex concepts and problems gradually.
In the domain of math problem-solving, it is natural to first learn the basic math operations and then progressively tackle more difficult problems. Therefore, we employ this strategy to guide the SFT data synthetic process. The stage 2 in the data synthesis pipeline is specifically designed for models to focus on mastering the more challenging problems. In this stage, we utilize the challenging problems, i.e., those categorized as Level 4 or Level 5 in the MATH dataset~\citep{hendrycks2021measuring} to generate additional 0.4 million query-response pairs. Finally, combined with 2.1M normal synthetic problems in stage 1, we obtain the 2.5M-instance Skywork-MathQA dataset. The rationale behind using these two stages and the experimental analysis of their impacts are discussed in Section~\ref{sec:level}.
We further fine-tune the intermediate models on these hard synthetic problems to obtain the Skywork-Math model series, which exhibit strong mathematical reasoning abilities.

\paragraph{Remark}
It is worth noting that the accuracy of our utilized GPT-4 version on the MATH benchmark is approximately 50\%, indicating that about half of our synthetic data in Skywork-MathQA dataset may contain minor errors in their results and intermediate reasoning process. However, scaling these SFT synthetic data reveals a clear positive trend in the performance of LLMs~(ref. Figure~\ref{scale_figure}). An interesting experimental phenomenon is that before reaching the upper bound performance of Skywork-Math 7B model series, data quantity seems to play a more important role than data quality.

%% file: sections/experiment.tex
\section{Experiment}

\subsection{Experimental Setup}
\subsubsection{Evaluation Datasets}
We primarily conduct our experiments on two benchmarks widely recognized for assessing mathematical reasoning capabilities. (1) \textbf{GSM8K}~\citep{cobbe2021trainingGSM8K} comprises a collection of high-quality math word problems at the grade school level. It contains 1,319 test questions. Typically, the reasoning steps in GSM8K vary between two and eight, ultimately yielding an integer as the answer.
(2) \textbf{MATH}~\citep{hendrycks2021measuring} contains 5,000 test questions, featuring math competition-level problems. The answers in GSM8K are integer, making it relatively easy for the regular expression matching program in evaluation frameworks to extract and verify answers. However the answers in MATH may contain complex mathematical formulas~(e.g., $\frac{2+\sqrt{2}}{4}$, $(\sqrt{2}, \sqrt{3})$). We have explored several evaluation benchmarks to assess the results on MATH~(e.g., \citep{yu2023metamath,simpleevals, shao2024deepseekmath, he2024olympiadbench}). Different evaluation benchmarks implement different regular expression rules to extract mathematical formulas, leading to significant performance variations among them~(in some cases, there are up to 5\% accuracy variations on MATH). 
In this paper, we adopt the same evaluation framework as in MetaMath~\citep{yu2023metamath} because it is widely used and provides strict and robust evaluation results using zero-shot and COT techniques.

\subsubsection{Pre-Trained Models}
We utilize three publicly available top-performing 7B pre-trained LLMs in the Skywork-MathQA models to push the limit of mathematical reasoning abilities in small-scale LLMs. 
Our empirical results indicate that Skywork-MathQA 7B models even outperform the recently released 70B LLaMA-3 Instruct Model~\citep{llama3modelcard} on the MATH benchmark.
\begin{itemize}
    \item \textbf{LLaMA2-7B}~\citep{touvron2023llama} is a general-purpose LLM model that has demonstrated significant performance across various benchmarks. However, it exhibits limited mathematical reasoning abilities.
    \item \textbf{Mistral-7B}~\citep{jiang2023mistral} is another general-purpose LLM model that exhibits strong reasoning abilities in math problem-solving and code generation.
    \item \textbf{DeepSeekMath-Base-7B}~\citep{shao2024deepseekmath} is a specialized LLM model tailored for mathematics reasoning. It stems from DeepSeek-Coder-Base-v1.5-7B~\citep{guo2024deepseekcoder} and has been further pre-trained on a mathematical corpus with 120 billion tokens. Due to this extended pre-training on massive math corpus, we observe a notable performance divergence between the specialized model and general-purpose LLM model~(ref. Section~\ref{sec:exp_scale}).
\end{itemize}

\subsubsection{Implementation Details}
We utilize the GPT-4 API with a temperature of $0.7$ to generate query-response pairs in Skywork-MathQA dataset.
To prevent data leakage, we evaluate the Skywork-Math models on the test examples of GSM8K and MATH with a 30-gram hit, as suggested by~\citep{azerbayev2024llemma}.
For all experiments, including ablations, Skywork-MathQA models are trained for $3$ epochs. A global batch size of $32$ is used along with the AdamW optimizer without weight decay. Following the original configurations of 7B pre-trained models, the learning rate is set to $2e-5$
for LLaMA2-7B and $2e-6$ for both Mistral-7B and DeekSeekMath-Base-7B. The learning rate warm-up ratio is $0.03$. All experiments are conducted on 8 Nvidia A800 GPUs with 80G memory. 
For evaluation, we use the vLLM~\citep{kwon2023efficientvLLM} library to generate inference responses, using the same prompt as in the SFT stage described in Section~\ref{sec:synthesis}.
Unless otherwise noted, we set the maximum length of models to 2048 in both the model SFT stage and the evaluation stage. 
We employ a stringent criterion similar to that used in Metamath~\citep{yu2023metamath}, achieving nearly 100\% precision but at the cost of a relatively low recall rate. This approach results in several instances where correct responses from the model are mistakenly labeled as incorrect according to our criteria. Specific examples can be found in Appendix~\ref{sp cases of ans}.

\subsection{Main Results}

\input{tables/table-1}
\subsubsection{Comprehensive Performance Comparison with State-of-the-art Models}
Table~\ref{main results} presents the comparison of Skywork-Math model series with the state-of-the-art closed- and open-source models on the test set of GSM8K and MATH benchmark to evaluate their math reasoning abilities. 
Because GPT-4-Turbo is a commercially closed-source model and cannot be fine-tuned to adhere to specific output formats, its responses are evaluated using a grading criterion with 4-shot COT prompting as used in~\citep{zheng2023progressivehint}.
(1) For the MATH benchmark, our Skywork-Math model series have achieved the state-of-the-art performance among LLM models smaller than 10B parameters with only the SFT technique, even surpassing the an early version of GPT-4. These results indicate that strong math reasoning abilities can be injected during the SFT stage through the extensive and high-quality Skywork-MathQA dataset. Moreover, Skywork-Math 7B models achieve competitive accuracy with 70B LLM models, which suggests 7B common LLM models can possess the strong math reasoning abilities with sufficient SFT process. These results demonstrate the significant effectiveness of our proposed two-stage data synthesis and model SFT pipeline.
(2) For the GSM8K benchmark, the Skywork-Math model series also achieve comparable performance with several state-of-the-art models. It is noteworthy that our Skywork-MathQA dataset contains no data referencing GSM8K. The characteristics of math word problem~(GSK8K) and math competition-level problems~(MATH) differ in their problem-answer formats and difficulty.
We posit that the success can be attributed to the difficulty of the relatively easy problems in MATH~(Level 1\&2) being similar to those in GSM8K, and the knowledge learned from solving competition-level mathematical problems can be effectively transferred to math word problems.

\subsubsection{Scaling Laws in SFT on Mathematical Reasoning}
\label{sec:exp_scale}
\begin{figure}[H]
    \centering
    \includegraphics[width=0.87\textwidth]{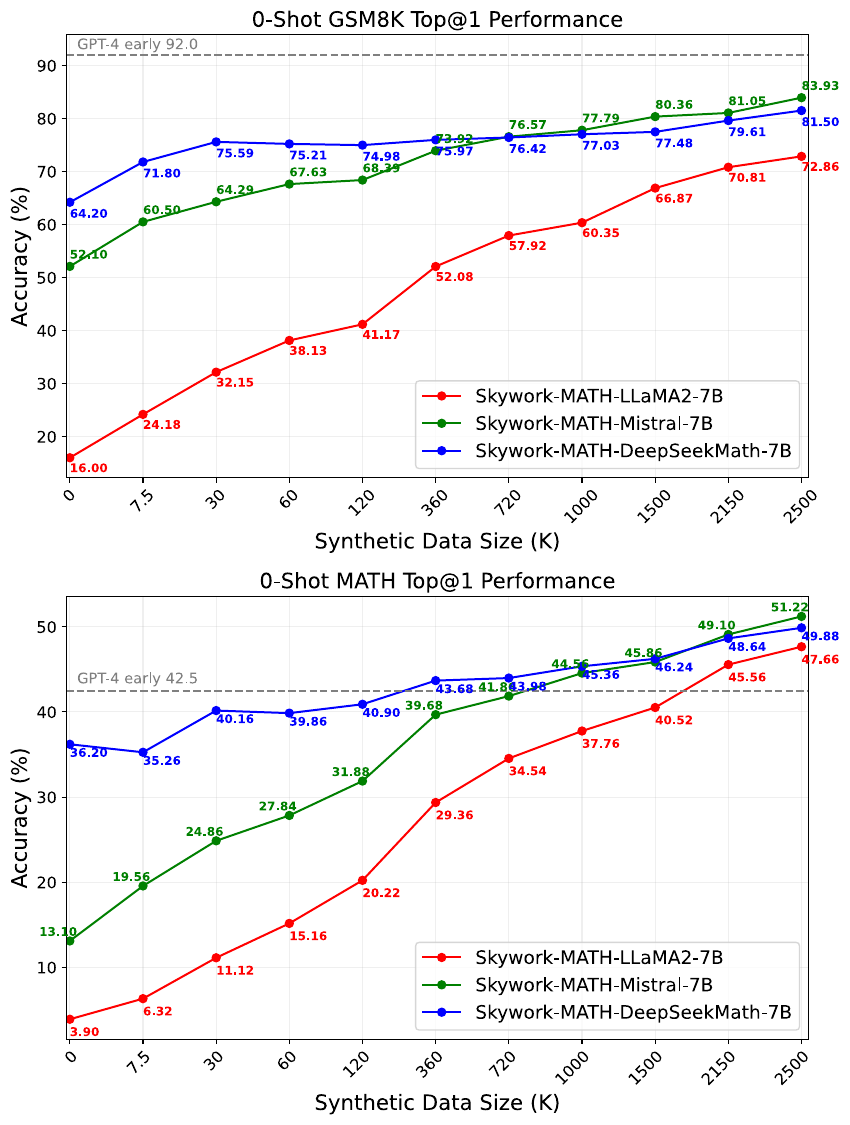}
    \vspace{-0.2cm}
    \caption{The zero-shot top1 performance of Skywork-Math 7B model series improves significantly as we scale up the size of synthetic SFT data in the Skywork-MathQA dataset. There is a clear trend indicating that the model’s math reasoning quality increases substantially with increases in data quantity.}
    \label{scale_figure}
    \vspace{-0.4cm}
\end{figure}

In Figure~\ref{scale_figure}, we illustrate the relationship between synthetic SFT dataset size and model performance on GSM8K and MATH. The curve clearly exhibits a scaling law relationship between the size of SFT data and model's performance. Here are some in-depth observations: 

\paragraph{Quantity Breeds Quality.}
To enhance the mathematical reasoning abilities in LLMs, increasing the quantity of synthetic data can significantly improve the quality of model performance. This scaling trend implies that, while SFT with a small amount of data could achieve decent results~\cite{zhou2023lima}, utilizing a larger scale of synthetic SFT data can further improve math reasoning performance.

\paragraph{Diminishing Returns from Continual Pre-Training.}
The DeepSeekMath-Base~\citep{shao2024deepseekmath} 7B model, which has been continually pre-trained with 120B math-related tokens sourced from the web, initially demonstrates superior performance. However, as we increase the synthetic dataset size in the Skywork-MathQA dataset, this advantage diminishes and is eventually surpassed by the Mistral~\citep{jiang2023mistral} 7B base model. As the amount of SFT data increases, Skywork-Math-Mistral-7B and Skywork-Math-LLaMA2-7B catch up in performance to the Skywork-Math-DeepSeekMath-7B. This suggests that while specialized pre-training provides a strong initial boost, its benefits are not consistently scalable and can be matched by increasing the quantity of synthetic SFT data.

\paragraph{Effect of Problem Difficulty.}
The accuracy performance for Skywork-Math 7B model series significantly increases as the synthetic data size expands from 2.1M to 2.5M, corresponding to the stage 2 in our data synthesis pipeline. This performance improvement in the final stage of data scaling indicates that incorporating more complex problems--- ranging from Level 3 to Level 5 in the MATH dataset---has a substantial positive impact on model performance. This finding underscores the importance of not only generating a large quantity of data but also including more challenging problems to push the limits of math reasoning abilities of LLM models. We will discuss this in more detail in Section~\ref{sec:level}.

\subsection{Experimental Analysis}
\input{tables/table-2}
\subsubsection{Fine-Grained Analysis across Different Difficulty Levels}
\label{sec:level}
We explore model's performance across various difficulty levels to analyze the internal relationship between data difficulty and LLM model's capability. The difficulty level distribution 
\begin{wrapfigure}{r}{0.5\textwidth}
    \centering
    \includegraphics[width=0.5\textwidth]{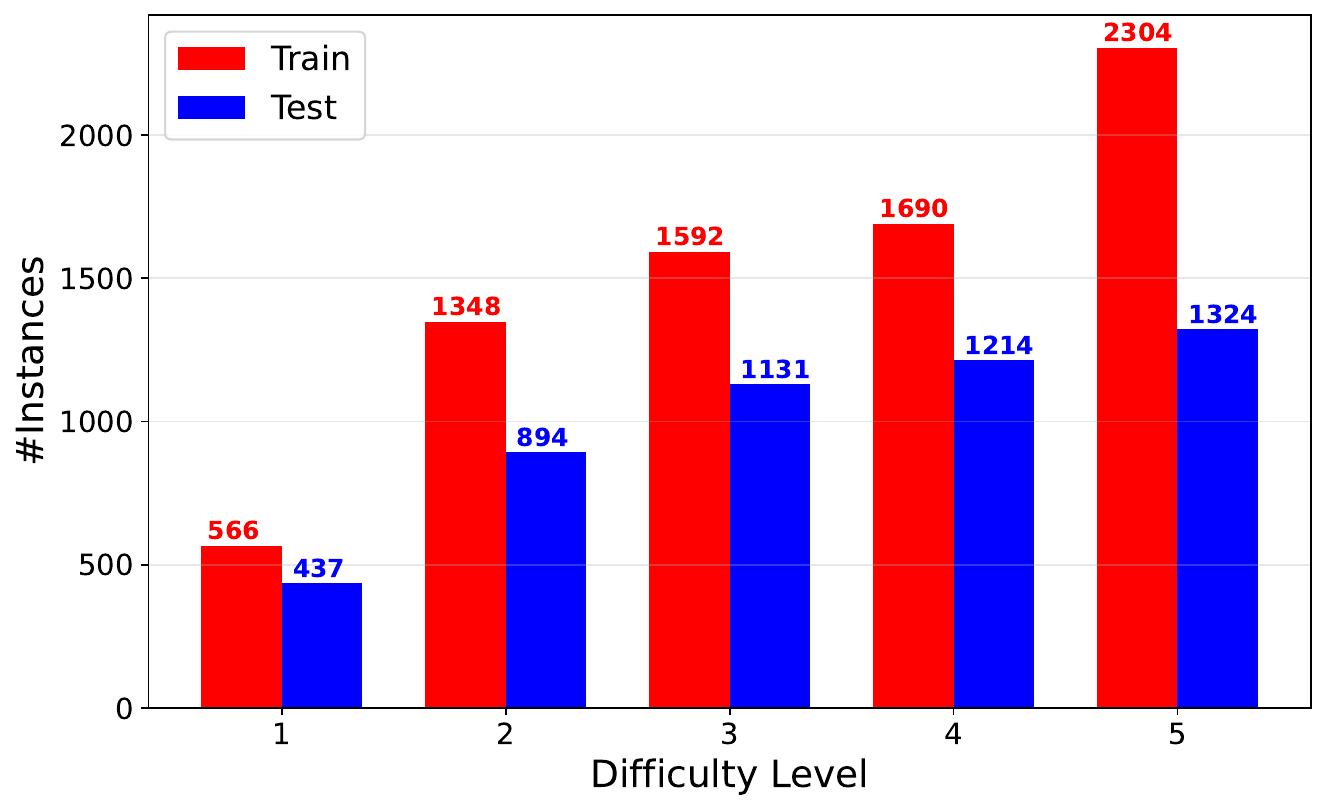}
    \caption{Difficulty level distribution of the training and test set in the MATH benchmark.}
    \label{math_dist}
\end{wrapfigure}
of the training and test set in MATH is illustrated in Figure~\ref{math_dist}. 
We can find that the number of hard problems~(Level 3-5) is much larger than that of easy problem~(Level 1\&2) in both training and test sets. This highlights the value of hard problems to improve the overall math reasoning performance.

In Table~\ref{level wise acc table}, we conduct comprehensive experiments with three pre-trained base LLM models in Skywork-Math 7B model series.
We observed a significant increase in accuracy for easy problems~(Level 1\&2) when scaling the dataset size from 7.5K to 2.1M, even reaching accuracies comparable to GPT-4-Turbo. However, the increase in accuracy for hard problems (Level 3-5) was relatively modest compared to GPT-4-Turbo.
This could be due to the lack of high-quality responses in hard problems, motivating us to perform the stage 2 in our data synthesis pipeline to generate hard synthetic problems. 
After fine-tuning our Skywork-Math 7B model with additional 0.4M hard synthetic problems, we observe a further increase in model performance, particularly at Level-3 and Level-4 on MATH.
For comparison, we conduct an experiment to fine-tune three base models in Skywork-Math 7B models using 0.4M hard synthetic problems along with the randomly sampled 7.5k problems. We notice that for hard problems~(Level 3-5), base models fine-tuned on the "2.1M + 0.4M~(hard)" data perform significantly better than those fine-tuned on the "7.5k + 0.4M (hard)" data. This supports the rationale that LLM models should acquire mathematical reasoning abilities progressively from easy to hard problems.
More detailed experiments can be found in Appendix~\ref{model perf on hard lv}. In addition to testing on different levels, we also conducted experiments on various math subjects, as detailed in Appendix~\ref{MATH Problems Study}.

\begin{figure}[t]
    \centering
    \includegraphics[width=\textwidth]{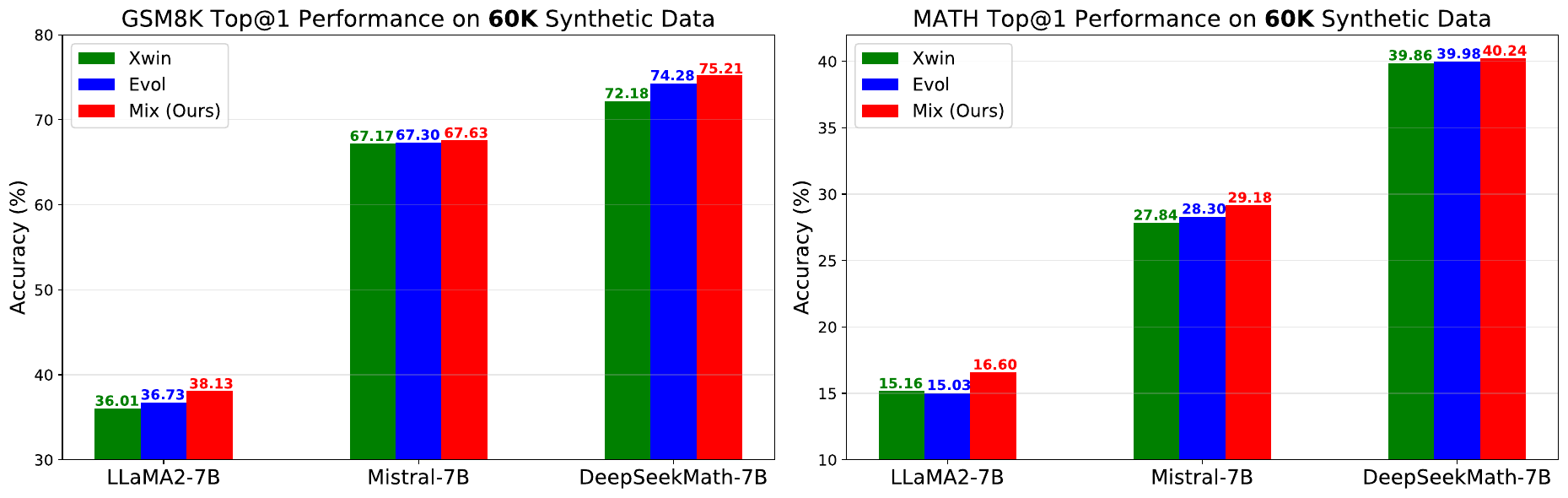}
    \caption{Performance of different base models in Skywork-Math 7B models with various data augmentation methods on GSM8K and MATH. "Mix" represents a combination of data generated by three augmentation methods detailed in Section~\ref{sec:data_augmentation}. For this ablation study, we utilize 60K synthetic SFT data in the Skywork-MathQA dataset.}
    \label{augment type diversity}
    \vspace{0.3cm}
\end{figure}

\subsubsection{Effect of Data Diversity}
\label{exp:diversity}
\paragraph{Diversity on Data Augmentation Methods.}
\label{exp:diversity:method}
One dimension of diversity is the data augmentation methods. We select 60K synthetic data in the Skywork-Math dataset to study this problem. As shown in Figure~\ref{augment type diversity}, the "Mix" approach, a combination of synthetic data generated by three augmentation methods, achieves the highest performance. Therefore, we utilize the "mix" method to generate our Skywork-MathQA dataset. Moreover, the Xwin-style~\citep{li2024common} approach and the MetaMathQA-style~\citep{yu2023metamath} approach require extensive time for answer verification and two steps for data generation, respectively. For the consideration of efficiency, we utilize the Evol-style~\citep{luo2023wizardmath} approach as a major component of the synthetic data due to requiring fewer input and output tokens within LLM models. 
We also observe that the impact of the mix rate of augmentation methods is not significant on the GSM8K and MATH benchmarks. However, combining these data augmentation methods is crucial for enhancing the data diversity of the Sykwork-MathQA dataset. Detailed exploration of data mixtures with different data augmentation methods is left for future work.

\input{tables/table-3}

\paragraph{Diversity of Seed Problems.}
Another dimension of diversity is the selection of seed problems. We construct two SFT datasets, each comprising 360K entries. The first dataset uses only the training set of MATH as the seed problems. The second dataset employs the diversity selection method introduced in Section~\ref{sec:diversity_selection}, which includes a wide range of non-proving problems from multiple academic data sources and uses the diversity selection method to further ensure the diversity.
As illustrated in Table~\ref{Diversity Table}, the improved diversity of seed problems in SFT data substantially enhances the math reasoning abilities in Skywork-Math models across three 7B base LLM models.

\input{tables/table-4}

\subsubsection{Data Selection with a Verifier}
Since the accuracy of GPT-4 on MATH is around $50\%$, we can infer that approximately half of the data samples in the Skywork-MathQA dataset may not have the right solving processes and answers. 
To ensure the collection of high-quality data, it is a natural way to perform data selection with a verifier to filter out wrong responses. 
We first eliminate synthetic data entries that fail to align with the ground truth final answers.
However, most data samples either lack the ground truth final answers or contain errors in intermediate reasoning steps. Therefore, we should design a more precise approach to ensure the entire solution is consistent with the ground truth.
We fine-tune a Mistral-7B~\citep{jiang2023mistral} base model with few-shot prompting to verify if the reasoning paths and final answers are correct.
Finally, we obtain approximately 1 million samples deemed correct by this fine-tuned verifier. With human verification of the results judged by the trained Mistral-7B verifier, it achieves an accuracy of approximately $80\%$. After implementing our filtering process, the fraction of correct data~($80\%$) increases significantly compared to its original fraction~($50\%$). As shown in Table~\ref{verifier filter 1M}, we present the results selected using the trained verifier in contrast to a random selection in the Skywork-Math dataset. 
We initially anticipated that, after filtering for correctness to obtain the 1M filtered dataset, the accuracies on GSM8K and MATH would range between 1M to 1.5M samples with random selection due to their quantitative relationship. \textit{However, the actual performance on the LLaMA2-7B and Mistral-7B models showed that the 1M filtered dataset performed even worse than the 1M dataset with random selection.} 

The experimental results align with the conclusion in DsDm~\citep{engstrom2024dsdm}. The data selection process on math reasoning is non-trivial and there exist multiple objectives to affect this data selection process. Our observation suggests that although the accuracy reaches as high as 80\%, the difficulty level of the selected problems significantly decreases. The selection process improves the data quality but significantly decreases the difficulty level of problems, thereby negatively impacting the performance of LLMs.
In order to filter out correct problems, the verifier model predominantly selects those problems with lower difficulty. 
To address the scarcity of hard problems in the filtered dataset, we further utilize GPT-4 with the COT prompt to pick out around 360K hard problems. 
Table~\ref{verifier filter composed360k} demonstrates that data selection with hard problems is effective, as all base models in the Skywork-Math models show improved performance on both the MATH and GSM8K benchmarks compared to their random selection counterparts.

\input{tables/table-5}

\input{tables/table-6}

\subsubsection{Can Math Reasoning Abilities Transfer Between Bilingual Language?}
The common view holds that mathematical problems mainly consist of symbols and expressions, and the textual language used to state them is not crucial for understanding. To explore whether math reasoning abilities can transfer between bilingual languages, we translate the GSM8K~\citep{cobbe2021trainingGSM8K} and MATH~\citep{hendrycks2021measuring} benchmarks from English to Chinese for bilingual language testing. It is important to note that all models are fine-tuned only on English data. 
As shown in Table~\ref{multi-lingual perf}, the overall math reasoning abilities are maintained between English and Chinese. There is a relatively small-scale performance degradation on MATH between English and Chinese, especially in Skywork-Math-Mistral-7B and Skywork-Math-DeepSeekMath-7B. However, there is a significant performance drop on GSM8K between English and Chinese, with up to a 20-point drop in Skywork-Math-LLaMA2-7B. Since GSM8K is grouped in the math word problem category, which requires more linguistic understanding, the degradation in accuracy is greater than that for MATH. Notably, Skywork-Math-DeepSeekMath-7B performs well in both English and Chinese. We hypothesize the reason for this is that the 120B continual pre-training corpus in the DeepSeekMath-Base-7B model includes many Chinese sources, which improves its Chinese language understanding. These results highlight the challenges associated with language dependence in understanding and performing mathematical reasoning tasks. 

\subsubsection{Can Math Reasoning Abilities Be Maintained in Robustness Tests?}
As suggested in CMATH~\citep{wei2023cmath}, several open-sourced LLM models, except GPT-4-Turbo, are vulnerable to robustness tests of math reasoning abilities influenced by distractors. To ascertain if models effectively comprehend the fundamental elements of mathematical word problems and their solutions, we inject each problem in GSM8K with 1-5 distractors as implemented in CMATH~\citep{wei2023cmath}. An example of two distractors is shown in Figure~\ref{cmath exp}. As listed in Table~\ref{cmath score}, open-sourced fine-tuned LLM models are sensitive to the distractors injected into math word problems. Compared to the MetaMathQA SFT dataset, our proposed Skywork-MathQA dataset significantly improves robustness performance in GSM8K based on common pre-trained models, such as Mistral-7B and DeepSeekMath-7B. 
We hypothesize that the reason lies in the significantly larger size of the Skywork-MathQA dataset compared to the MetaMathQA dataset. The improved diversity of the Skywork-MathQA dataset can help the LLM models STF on it to better withstand robustness tests.
However, GPT-4-Turbo consistently excludes interference information and focuses on the relevant information, thereby producing correct responses with even 5 distractors in GSM8K. These results suggest that most of open-source SFT models cannot truly understand the semantic information of math world problems but rather mechanically extract numbers from the sentence and calculate them. Effectively improving math reasoning abilities while maintaining robustness like GPT-4-Turbo is an important area for future exploration.

\begin{figure}[t]
\begin{tcolorbox}[colback=blue!3!white, left=6pt, right=0pt]
\textbf{Original Question}: There were a total of 15 fish in the plate. After the kitten ate some, there were 10 fish left. How many fish did the kitten eat?

\hdashrule[0.5ex]{\linewidth}{1pt}{3pt}

\textbf{Question with Distractors}: There are 3 kittens in the house. There were a total of 15 fish in the plate, including 10 carp and 5 belt fish. After the kittens ate some fish, there were still 10 fish left. How many fish did the kittens eat?

\hdashrule[0.5ex]{\linewidth}{1pt}{3pt}

\textbf{Distractors:}
\vspace{-0.5em}
\begin{itemize}
    \item[1.] "There are 3 kittens in the house."
    \item[2.] "Including 10 carp and 5 belt fish."
\end{itemize}
\end{tcolorbox}
\vspace{-1em}
\caption{An example of an original question from GSM8K~\citep{cobbe2021trainingGSM8K} and the same question with two distrators as implemented in CMATH~\citep{wei2023cmath}.}
\label{cmath exp}
\end{figure}

\input{tables/table-7}

\subsubsection{Ablation Studies Between Sparse MOE and Dense Models}
Recent advancements have witnessed the rapid development of sparse MOE models~\citep{deepseekv2}.
To evaluate the generalization capability of our Skywork-MathQA dataset across both sparse MOE and dense models, we select commonly used dense~(Skywork-Math-Mistral-7B~\citep{jiang2023mistral}) and sparse MOE~(Mixtral-8$\times$7B~\citep{jiang2024mixtral}) models as the pre-trained LLM base models. We conduct experiments using the Skywork-MathQA dataset in both stage 1 and stage 2. As shown in Table~\ref{table:moe}, the results confirm strong generalization across different types of LLM models. 
However, the Mixtral-8$\times$7B fine-tuned on the Skywork-MathQA dataset does not show superior performance compared with its dense counterpart. The Mixtral-8$\times$7B and Skywork-Math-Mistral-7B almost exhibit almost identical performance on GSM8K and MATH. We posit the reason is that the sparse MoE model, due to its mixture-of-expert architecture, may not significantly improve the performance on the specific task~(i.e., the math reasoning task), but can better handle task-specific knowledge without compromising performance on other tasks~\citep{xue2024openmoe, wei2024skywork}.

\input{tables/table-8}

\begin{figure}[t]
\begin{tcolorbox}[colback=blue!3!white, left=6pt, right=0pt]

\textbf{Question in Skywork-MathQA}: \textcolor{red}{Let $x$ and $y$ be nonzero real numbers such that} \[(3-4i)(x+yi)\] is pure imaginary. Find $x/y$.

\hdashrule[0.5ex]{\linewidth}{1pt}{3pt}

\textbf{Question in the MATH test set}: \textcolor{red}{Let $x$ and $y$ be nonzero real numbers such that} \[xy(x^2-y^2) = x^ 2 + y^2.\] Find the minimum value of $x^2 + y^2$.

\end{tcolorbox}
\vspace{-1em}
\caption{An example of the math questions that are completely different but get filtered by a 10-gram filter due to a common condition.}
\label{fig:iqc_example}
\end{figure}

\input{tables/table-9}

\subsubsection{Effect of Data Leakage}
Though we never use the test data from MATH~\citep{hendrycks2021measuring} or GSM8K~\citep{cobbe2021trainingGSM8K} for fine-tuning LLM models, we utilize GPT-4~\citep{openai2024gpt4} to synthesize data, which may inadvertently contaminate our synthetic dataset with elements from the test data in the evaluation benchmarks. Therefore, we follow a standard 30-gram filtering process~\citep{azerbayev2024llemma} on test data of the corresponding benchmark to circumvent the data leakage of the Skywork-MathQA dataset. We filter out approximately 6K samples for the test set of MATH and none for GSM8K.

To assess the impact of the n-gram filter, we tested a stricter 10-gram filter, which is much more stringent than the 30-gram filter. We observe that the 10-gram filter removes a lot of data that has little relation to the data in the test set of MATH. As illustrated in Figure~\ref{fig:iqc_example}, there are two entirely unrelated examples in our synthetic Skywork-MathQA dataset and the test set of MATH. It is evident that "Let $x$ and $y$ be nonzero real numbers such that" is a very common condition in math problems. The 10-gram filter results in the removal of many completely unrelated problems in the synthetic data. Consequently, we use the 30-gram filter instead of the 10-gram filter to produce the Skywork-MathQA dataset.

We further conduct experiments to quantitatively analyze the difference between the 30-gram and 10-gram filter using our Skywork-MathQA dataset in stage 1. 
Our Skywork-MathQA dataset, which has already been filtered using the 30-gram filter, consists of 2.16M instances. After applying 10-gram filtering, we have 2.10M instances. The filtered-out data, meaning the data samples present in the 2.16 million instances but not in the 2.10 million instances, consists of 60K samples. For a fair comparison, we also randomly select 60K data samples from the Skywork-MathQA dataset. 
The results of accuracies on the MATH benchmark are reported in Table~\ref{detailed n-gram perf}. The observations are as follows: (1) The 10-gram filter is too strict, leading to the removal of some specific types of problems in the math benchmark~(ref. Figure~\ref{fig:iqc_example}), which results in performance degradation.
(2) The 60K randomly sampled data is much more useful than the 60K filtered-out data for Skywork-Math-LLaMA2-7B and Skywork-Math-Mistral-7B. The experimental results are reasonable, as the diversity in the randomly selected 60K data is much greater than that in the filtered 60K data.
(3) The performance of DeepSeekMath-7B after SFT with the 2.10M dataset is significantly worse than with the 2.16M dataset. The filtered 60K dataset performs even better than the randomly selected 60K dataset. We believe this is because Skywork-Math-DeepSeekMath-7B may focus on the types of problems present in the filtered 60K data. Its base model, DeepSeekMath-Base-7B~\citep{shao2024deepseekmath}, is a specialized math LLM model continually pre-trained on a large collection of math data that matches some of the types in these filtered 60K problems.

\input{tables/table-10}

\subsubsection{Effect of Model Maximum Length}
As the difficulty level of problems increases, the length of reasoning steps typically becomes longer, especially with those generated by LLMs. If the model's maximum length is too small, the response may be truncated. In our synthetic Skywork-MathQA SFT dataset, around 130K problems exceed 512 tokens. Therefore, we set the maximum length of models to 2048 tokens in both the SFT stage and the evaluation stage. As shown in Table~\ref{table:length}, increasing the model's maximum length leads to improved performance, indicating that 7B models can comprehend and execute long reasoning processes.

%% file: tables/table-1.tex
\begin{table*}[htbp]
\centering
\scalebox{0.9}{
\begin{small}
\begin{tabular}{lccc}
\toprule
\textbf{Model} & \textbf{\#Params} & \textbf{GSM8K(\%)} & \textbf{MATH(\%)} \\ 
\midrule
\midrule
\multicolumn{4}{c}{Closed-source models} \\
\midrule
GPT-3.5-Turbo \citep{gpt3.5} & N/A & 80.8 & 34.1 \\
GPT-4-Turbo \citep{openai2024gpt4} & N/A & 90.51 & 57.0 \\
GPT-4  \citep{openai2024gpt4} & N/A & 92.0 & 42.5 \\
PaLM2 \citep{anil2023palm2} & 540B & 80.7 & 34.3 \\
Flan-PaLM2 \citep{anil2023palm2} & 540B & 84.7 & 33.2 \\
Minerva \citep{lewkowycz2022minerva} & 8B & 16.2 & 18.1 \\
Minerva \citep{lewkowycz2022minerva} & 62B & 52.4 & 27.6 \\
Minerva \citep{lewkowycz2022minerva} & 540B & 58.8 & 33.6 \\
ChatGLM3-32B-SFT-2312 \citep{xu2024chatglmmath} & 32B & 75.8 & 29.0 \\
~~~~~~~~~+RFT, DPO \citep{xu2024chatglmmath} & 32B & 82.6 & 40.6 \\
Claude-3-Oppus \citep{claude3modelcard} & N/A & \textbf{95.0} & \textbf{60.1}\\

\midrule
\multicolumn{4}{c}{Open-source models (1-10B)} \\
\midrule
Baichuan-2 \citep{yang2023baichuan2} & 7B & 24.5 & 5.6 \\
LEMA-LLaMA2 \citep{an2024learningLEMA} & 7B & 54.1 & 9.4 \\
MetaMath  \citep{yu2023metamath} & 7B & 66.5 & 19.8 \\
WizardMath-V1.1  \citep{luo2023wizardmath} & 7B & 83.2 & 33.0 \\
Xwin-Math-LLaMA \citep{li2024common} & 7B & 82.6 & 40.6 \\
Xwin-Math-Mistral \citep{li2024common} & 7B & \textbf{89.2} & 43.7 \\
Xwin-Math-Llemma \citep{li2024common} & 7B & 84.2 & 47.2 \\
MAmmoTH \citep{yue2023mammoth} & 7B & 53.6 & 31.5 \\
InternLM2-Math \citep{ying2024internlm} & 7B & 78.1 & 34.6 \\
DeepSeekMath-Instruct \citep{shao2024deepseekmath} & 7B & 82.9 & 46.8 \\
\rowcolor{greyrow} Skywork-Math-LLaMA2 (ours) & 7B & 72.9 & 47.7 \\
\rowcolor{greyrow} Skywork-Math-Mistral (ours) & 7B & 83.9 & \textbf{51.2} \\
\rowcolor{greyrow} Skywork-Math-DeepSeekMath (ours) & 7B & 81.5 & 49.9 \\
LLaMA3-Instruct \citep{llama3modelcard} & 8B & 79.6 & 30.0 \\

\midrule
\multicolumn{4}{c}{Open-source models (10-50B)} \\
\midrule
LLaMA2 \cite{touvron2023llama} & 13B & 28.70 & 3.90 \\
Baichuan-2 \citep{yang2023baichuan2} & 13B & 52.8 & 10.1 \\
MetaMath  \citep{yu2023metamath} & 13B & 72.3 & 22.4 \\
Wizard-Math \citep{luo2023wizardmath} & 13B & 63.9 & 14.0 \\
MAmmoTH \citep{yue2023mammoth} & 13B & 62.0 & 34.2 \\
LEMA-LLaMA2 \citep{an2024learningLEMA} & 13B & 65.7 & 12.6 \\
Xwin-Math \citep{li2024common} & 13B & \textbf{88.1} & \textbf{44.9} \\
InternLM2-Math \citep{ying2024internlm} & 20B & 82.6 & 37.7 \\
LLaMA2 \cite{touvron2023llama} & 34B & 42.20 & 6.20 \\
LLema \citep{azerbayev2024llemma} & 34B & 51.5 & 25.0 \\

\midrule
\multicolumn{4}{c}{Open-source models (50-70B)} \\
\midrule
WizardMath \citep{luo2023wizardmath} & 70B & 81.6 & 22.7 \\
MetaMath \citep{yu2023metamath} & 70B & 82.3 & 22.6 \\
LLaMA2 \citep{touvron2023llama} & 70B & 56.8 & 13.5 \\
LEMA-LLaMA2 \citep{an2024learningLEMA} & 70B & 83.5 & 25.0 \\
MAmmoTH \citep{yue2023mammoth} & 70B & 76.9 & 41.8 \\
LLaMA3-Instruct \citep{llama3modelcard} & 70B & 90.0 & 50.4 \\
Xwin-Math \citep{li2024common} & 70B & \textbf{90.6} & \textbf{52.8} \\

\bottomrule
\end{tabular}
\end{small}
}
\caption{Summary of math reasoning performance of closed- and open-source LLM models in terms of accuracy~(\%). All results for open-source models are reported as top1 accuracy using only SFT techniques. Skywork-Math models employ zero-shot chain-of-thought~(COT) evaluation framework as implemented in MetaMath~\citep{yu2023metamath}. The best result in each block are highlighted in bold. GPT-4-Turbo is evaluated using the grading criteria with 4-shot COT prompting as implemented in~\citep{zheng2023progressivehint}. Skywork-Math 7B models, using only synthetic SFT
data, have achieved SOTA performance on MATH among models small than 10B parameters, even outperforming 70B LLM models and an early version of GPT-4.}
\label{main results}
\end{table*}

%% file: tables/table-2.tex
\begingroup
\setlength{\tabcolsep}{3pt} 
\renewcommand{\arraystretch}{1} 
\begin{table*}[t]
\centering
\adjustbox{max width=\textwidth}{
\begin{tabular}{lcccccc}
\toprule
\multirow{2}{*}{\textbf{Base Model}} & \multirow{2}{*}{\textbf{{Dataset Size}}} & \multicolumn{5}{c}{\textbf{Difficulty Levels in MATH(\%)}} \\
\cmidrule(l){3-7}
& & \textbf{Level-1} & \textbf{Level-2} & \textbf{Level-3} & \textbf{Level-4} & \textbf{Level-5} \\
\midrule
LLaMA2-7B & 7.5K       & 17.85 & 8.39  & 4.77  & 3.05  & 0.91 \\
Mistral-7B & 7.5K      & 37.99 & 25.17 & 15.12 & 8.48  & 2.49 \\
DeepSeekMath-7B & 7.5K & 64.07 & 46.76 & 37.84 & 24.63 & 10.73 \\
\midrule
LLaMA2-7B & 2.1M      & 78.03 & 60.29 & 48.19 & 35.09 & 19.56 \\
Mistral-7B & 2.1M     & 80.78 & 66.33 & 55.53 & 41.52 & 21.45 \\
DeepSeekMath-7B & 2.1M & 80.78 & 65.21 & 58.00 & 41.60 & 21.83 \\
\midrule
LLaMA2-7B & 7.5k + 0.4M~(hard) & 63.16 & 43.96 & 34.39 & 24.46 & 10.20 \\
Mistral-7B & 7.5k + 0.4M~(hard) & 71.62 & 57.27 & 48.72 & 34.60 & 16.99 \\
DeepSeekMath-7B & 7.5k + 0.4M~(hard) & 81.01 & 61.97 & 51.90 & 37.07 & 18.05 \\
\midrule
LLaMA2-7B & 2.1M + 0.4M~(hard) & 78.03 & 62.42 & 52.87 & 37.48 & 18.73 \\
Mistral-7B & 2.1M + 0.4M~(hard) & 83.52 & 67.56 & 60.65 & 44.89 & 25.08 \\
DeepSeekMath-7B & 2.1M + 0.4M~(hard) & 82.84 & 67.23 & 58.71 & 42.01 & 21.30 \\
\midrule
GPT-4-Turbo & - & 82.84 & 73.38 & 65.34 & 52.88 & 34.06 \\
\bottomrule
\end{tabular}
}
\caption{Accuracies~(\%) across difficulty levels~(from Level-1 to Level-5) with three base models in Skywork-Math 7B model series before and after fine-tuning on stage 2 in the MATH benchmark. 7.5K data samples are randomly sampled from the Skywork-MathQA dataset. GPT-4-Turbo is evaluated using our designed grading criteria with 4-shot COT prompting. In stage 1, Skywork-Math 7B models significantly improve the performance on easy problems in MATH~(Level 1\&2) using 2.1M synthetic SFT data. In stage 2, Skywork-Math 7B models show significant improvements on hard problems in MATH~(Level 3-5) using 2.5M synthetic SFT data.}
\label{level wise acc table}
\end{table*}
\endgroup

%% file: tables/table-3.tex
\begin{table}[t]
\centering
\scalebox{1.}{
\begin{tabular}{lccc}
\toprule
\textbf{Base Model} & \textbf{Diversity Selection}&\textbf{MATH(\%)} &\textbf{GSM8K(\%)} \\
\midrule
LLaMA2-7B & \ding{55} & 29.48 & 50.57\\
Mistral-7B & \ding{55} & 38.50 & 72.71\\
DeepSeekMath-7B & \ding{55} & 43.96 & 74.30\\
\midrule
LLaMA2-7B & \ding{51} & 29.36 & 52.08 \\
Mistral-7B & \ding{51} & 39.68 & 73.92 \\
DeepSeekMath-7B & \ding{51} & 43.68 & 75.97\\
\bottomrule
\end{tabular}
}
\caption{Ablation studies with the diversity selection method on 360K data samples applied in stage 1 of the data synthesis pipeline. \ding{51}~(\ding{55}) means that we evaluate w~(w/o) the diversity selection method.}
\label{Diversity Table}
\end{table}

%% file: tables/table-4.tex
\begingroup
\setlength{\tabcolsep}{3pt} 
\renewcommand{\arraystretch}{1} 
\begin{table*}[t]
\centering
\adjustbox{max width=\textwidth}{
\begin{tabular}{lccc}
\toprule
\textbf{Base Model} & \textbf{Dataset~(Size)} & \textbf{GSM8K(\%)} & \textbf{MATH(\%)} \\
\midrule
LLaMA2-7B & Random selection (1M) & 60.35 & 37.76 \\
LLaMA2-7B & Random selection (1.5M) & 66.87 & 40.52 \\
LLaMA2-7B & Selection with a verifier (1M) & 62.77 & 36.40 \\
\midrule
Mistral-7B & Random selection (1M) & 77.79 & 44.56 \\
Mistral-7B & Random selection (1.5M) & 80.36 & 45.86 \\
Mistral-7B & Selection with a verifier (1M) & 77.26 & 43.04 \\
\bottomrule
\end{tabular}
}
\caption{Comparisons of the model performance on GSM8K and MATH in terms of accuracy using random selection and selection with a verifier. All data samples are selected from the Skywork-MathQA dataset. Random selection on the math reasoning dataset is a simple but hard-to-beat strategy. Without a carefully designed filtering strategy, it is non-trivial to outperform random selection.}
\label{verifier filter 1M}
\end{table*}
\endgroup

%% file: tables/table-5.tex
\begingroup
\setlength{\tabcolsep}{3pt} 
\renewcommand{\arraystretch}{1} 
\begin{table*}[t]
\centering
\adjustbox{max width=\textwidth}{
\begin{tabular}{lccc}
\toprule
\textbf{Base Model} & \textbf{Dataset~(Size)} & \textbf{GSM8k(\%)} & \textbf{MATH(\%)} \\
\midrule
LLaMA2-7B & Random selection (360K) & 52.08 & 29.36 \\
LLaMA2-7B & Selection with hard problems (360K) & 54.36 & 36.68 \\
\midrule
Mistral-7B & Random selection (360K) & 73.92 & 39.68 \\
Mistral-7B & Selection with hard problems (360K) & 76.42 & 40.20 \\
\midrule
DeepSeekMath-7B & Random selection (360K) & 75.97 & 43.68 \\
DeepSeekMath-7B & Selection with hard problems (360K) & 75.74 & 44.48 \\
\bottomrule
\end{tabular}
}
\caption{Comparisons of the model performance on GSM8K and MATH in terms of accuracy using random selection and our designed selection strategy with filtering for more hard problems. All data samples are selected from the Skywork-MathQA dataset. Our strategy consistently outperform random selection.}
\label{verifier filter composed360k}
\end{table*}
\endgroup

%% file: tables/table-6.tex
\begingroup
\setlength{\tabcolsep}{3pt} 
\renewcommand{\arraystretch}{1} 
\begin{table*}[t]
\centering
\adjustbox{max width=\textwidth}{
\begin{tabular}{lcccc}
\toprule
\multicolumn{1}{l}{\multirow{2}{*}{\textbf{Model}}} & \multicolumn{2}{c}{\textbf{GSM8K(\%)}} & \multicolumn{2}{c}{\textbf{MATH(\%)}} \\
\cmidrule(r){2-3} \cmidrule(r){4-5}
        & \textbf{English} & \textbf{Chinese} & \textbf{English} & \textbf{Chinese}\\
\midrule
LLaMA3-8B + Skywork-MathQA & 75.97 & 58.83 & 50.30 & 44.10 \\ 
Mixtral-8x7B + Skywork-MathQA & 83.93 & 72.71 & 51.40 & 48.02 \\
Llemma-7B + Skywork-MathQA & 66.03 & 50.72 & 40.08 & 37.42 \\
Skywork-Math-LLaMA2-7B & 72.86 & 50.34 & 47.66 & 38.38\\
Skywork-Math-Mistral-7B & 83.93 & 69.75 & 51.22 & 48.34 \\
Skywork-Math-DeepSeekMath-7B & 81.50 & 73.69 & 49.88 & 48.22 \\
\bottomrule
\end{tabular}
}
\caption{Results of bilingual language testing on GSM8K and MATH. Note that all models are fine-tuned on English data. The Chinese version of GSM8K and MATH are translated from their English counterparts using GPT-4. LLaMA3-8B, Mixtral-8x7B, Llemma-7B are fine-tuned on our Skywork-MathQA datasets. Our empirical results indicates that the strong math reasoning capabilities can be maintained between English and Chinese.}
\label{multi-lingual perf}
\end{table*}
\endgroup

%% file: tables/table-7.tex
\begingroup
\setlength{\tabcolsep}{3pt} 
\renewcommand{\arraystretch}{1} 
\begin{table*}[t]
    \centering
    \adjustbox{max width=\textwidth}{
    \begin{tabular}{lccccccc}
        \toprule
        \multirow{2}{*}{\textbf{Model}} & \multirow{2}{*}{\textbf{SFT Dataset~(Size)}} & \multirow{2}{*}{\textbf{GSM8K(\%)}} & \multicolumn{5}{c}{\textbf{\#Distractors in GSM8K}} \\
        \cmidrule(l){4-8}
        & & & 1 & 2 & 3 & 4 & 5 \\
        \midrule
        GPT-4-Turbo & - & 90.51 & 95.30 & 91.44 & 88.98 & 88.02 & 85.37 \\
        DeepSeekMath-7B-Instruct & - & 82.90 & 73.77 & 62.97 & 51.44 & 48.22 & 43.88 \\
        \midrule
        Mistral-7B & MetaMathQA (395K) & 79.08 & 70.10 & 56.80 & 48.95 & 46.01 & 38.51 \\
        DeepSeekMath-7B & MetaMathQA (395K) & 82.49 & 73.20 & 60.33 & 50.26 & 42.31 & 39.40 \\
        LLaMA2-13B & MetaMathQA (395K) & 70.96 & 65.86 & 50.25 & 41.21 & 33.73 & 31.64 \\
        \midrule
        Llemma-7B & Skywork-MathQA (2.5M)  & 66.03 & 61.40 & 52.90 & 46.06 & 40.38 & 38.21 \\
        LLaMA3-8B & Skywork-MathQA (2.5M) & 75.97 & 75.14 & 70.91 & 65.35 & 62.43 & 55.82 \\
        Mixtral-8x7B & Skywork-MathQA (2.5M) & 83.93 & 84.19 & 78.21 & 73.36 & 68.93 & 66.57 \\
        Skywork-Math-LLaMA2-7B & Skywork-MathQA (2.5M) & 72.86 & 64.72 & 58.56 & 54.20 & 49.41 & 44.63 \\
        Skywork-Math-Mistral-7B & Skywork-MathQA (2.5M) & 83.93 & 83.16 & 75.19 & 72.57 & 66.42 & 67.01 \\
        Skywork-Math-DeepSeekMath-7B & Skywork-MathQA (2.5M) & 81.50 & 78.35 & 72.54 & 64.70 & 59.17 & 57.31 \\
        \bottomrule
    \end{tabular}
    }
    \caption{Performance against the number of the distractors added to the original GSM8K dataset. GPT-4 demonstrate remarkable robustness, while other models fail.}
    \label{cmath score}
\end{table*}
\endgroup

%% file: tables/table-8.tex
\begingroup
\setlength{\tabcolsep}{3pt} 
\renewcommand{\arraystretch}{1} 
\begin{table*}[t]
\centering
\adjustbox{max width=\textwidth}{
\begin{tabular}{lccc}
\toprule
\textbf{Model} & \textbf{Data Synthesis Pipeline~(Size)} & \textbf{GSM8K(\%)} & \textbf{MATH(\%)}\\
\midrule
Mistral-7B & - & 50.00 & 12.70 \\
Skywork-Math-Mistral-7B & Stage 1 (2.1M) & 83.25 & 49.10 \\
Skywork-Math-Mistral-7B & Stage 2 (2.5M) & 83.93 & 51.22 \\
\midrule
Mixtral-8$\times$7B & - & 74.40 & 28.40 \\
Mixtral-8$\times$7B +  Skywork-MathQA  & Stage 1 (2.1M) & 85.06 & 50.02 \\
Mixtral-8$\times$7B +  Skywork-MathQA & Stage 2 (2.5M) & 83.93 & 51.40 \\
\bottomrule
\end{tabular}
}
\caption{Performance comparison between the dense~(Skywork-Math-Mistral-7B) and sparse MOE~(Mixtral-8$\times$7B) LLM model. We fine-tune the corresponding base models using the Skywork-MathQA dataset in both stage 1 and stage 2 of the data synthesis pipeline.}
\label{table:moe}
\end{table*}
\endgroup

%% file: tables/table-9.tex
\begingroup
\setlength{\tabcolsep}{3pt} 
\renewcommand{\arraystretch}{1} 
\begin{table}[h]
\centering
\adjustbox{max width=\textwidth}{
\begin{tabular}{lcc}
\toprule
\textbf{Model} & \textbf{Filter Method~(size)} & \textbf{MATH(\%)} \\
\midrule
Skywork-Math-LLaMA2-7B & 30-gram (2.16M) & 45.56 \\
Skywork-Math-LLaMA2-7B & 10-gram (2.10M) & 37.54 \\
Skywork-Math-LLaMA2-7B & Filter-out (60K) & 10.76 \\
Skywork-Math-LLaMA2-7B & Random selection (60K) & 15.16 \\
\midrule
Skywork-Math-Mistral-7B & 30-gram (2.16M) & 49.10 \\
Skywork-Math-Mistral-7B & 10-gram (2.10M)  & 40.78 \\
Skywork-Math-Mistral-7B & Filter-out (60K) & 22.32 \\
Skywork-Math-Mistral-7B & Random selection (60K) & 27.84 \\
\midrule
Skywork-Math-DeepSeekMath-7B & 30-gram (2.16M) & 48.64 \\
Skywork-Math-DeepSeekMath-7B & 10-gram (2.10M) & 36.68 \\
Skywork-Math-DeepSeekMath-7B & Filter-out (60K) & 40.64 \\
Skywork-Math-DeepSeekMath-7B & Random selection (60K) & 39.86 \\
\bottomrule
\end{tabular}
}
\caption{Accuracies~(\%) on MATH for the Skywork-Math models using the 30-gram and 10-gram filter methods. "Filter-out" indicates samples present in the 30-gram filter method but not in the 10-gram filter method. For a fair comparison, we also randomly sampled 60K data points from our Skywork-MathQA dataset.
}
\label{detailed n-gram perf}
\end{table}
\endgroup

%% file: tables/table-10.tex
\begingroup
\setlength{\tabcolsep}{3pt} 
\renewcommand{\arraystretch}{1} 
\begin{table}[t]
\centering
\adjustbox{max width=\textwidth}{
\begin{tabular}{lccc}
\toprule
\textbf{Model} & \textbf{Model Maximum Length} & \textbf{MATH(\%)} & \textbf{GSM8k(\%)} \\
\midrule
Skywork-Math-LLaMA2-7B & 512  & 44.06 & 67.85 \\
Skywork-Math-LLaMA2-7B & 2048 & 47.66 & 72.86 \\
\midrule
Skywork-Math-Mistral-7B & 512 & 50.56 & 82.41 \\
Skywork-Math-Mistral-7B & 2048 & 51.22 & 83.93 \\
\midrule
Skywork-Math-DeepSeekMath-7B & 512 & 48.28 & 80.52 \\
Skywork-Math-DeepSeekMath-7B & 2048 & 49.88 & 81.50 \\
\bottomrule
\end{tabular}
}
\caption{Comparison of performance in Skywork-Math models using the 2.5M-instacne Skywork-MathQA dataset with different maximum model lengths.}
\label{table:length}
\end{table}
\endgroup

%% file: sections/conclusion.tex
\section{Closing Remarks and Future Directions}
We study how to empower mathematical reasoning abilities for common 7B pre-trained LLM models. We propose the Skywork-MathQA dataset, consisting of 2.5 million diverse and high-quality SFT instances, implemented through our novel two-stage data synthesis pipeline. We introduce Skywork-Math model series, demonstrating that common small-scale 7B language models can stimulate strong mathematical reasoning ability using only synthetic SFT data. Skywork-Math models achieve state-of-the-art accuracy among models smaller than 10B parameters using only synthetic SFT data, surpassing 70B LLM models and an early version of GPT-4 on MATH. These results suggest that the data scaling law for mathematical reasoning in LLM models remains significant and promising. Notably, this research provides several valuable insights and practical takeaways to advance our understanding of the capabilities and limitations of LLMs in mathematical reasoning.

Finally, we present two promising future directions for this work:
\vspace{-1em}
\paragraph{Code-Integrated Math Reasoning.} Complex scientific calculations are essential for tackling difficult mathematical problems. By embedding executable code, LLMs can dynamically generate and execute code to solve intricate mathematical problems, ensuring higher accuracy and robustness. Some recent works have already been proposed to translate mathematical problems into executable code~\citep{gou2023tora,toshniwal2024openmathinstruct1}. However, code cannot always be generated correctly on the first attempt. Therefore, iteratively utilizing code to solve challenging math problems is a promising direction for future research.
\vspace{-1em}
\paragraph{More General Reasoning Tasks.} Reasoning is a crucial ability for complex problem-solving. Beyond mathematical reasoning, there are many other important reasoning tasks, such as logical reasoning, causal reasoning, and commonsense reasoning~\citep{sun2023survey}. It is intriguing to explore how our proposed method can be applied to these more general reasoning tasks.

%% file: sections/acknowledgement.tex
\section{Acknowledgements}
We would like to thank Longhui Yu~(the author of MetaMath) and Chen Li~(the author of Xwin-Math) for their valuable discussions. Our deepest gratitude goes to our boss, Yahui Zhou, whose financial assistance in scaling supervised fine-tuning data and providing access to GPU computational resources was indispensable for the successful completion of this study. 

%% file: sections/appendix.tex
\section{Illustrations of Three Different Data Augmentation Methods}
\label{appendix:aug_exp}
We present three specific examples using the corresponding augmentation styles introduced in Section~\ref{sec:data_augmentation}. We use the same query to investigate the differences in the response. Overall, the differences among these three methods are nuanced, but combining them is crucial to enhance the diversity of the Skywork-MathQA dataset~(ref. Section~\ref{exp:diversity:method}).
In Figure~\ref{A.1}, the MetaMathQA-style data is answer-focused and maintains a coherent solving process. 
Figure~\ref{A.2} illustrates the Evol-style data, which provides a more detailed solution and includes extensive text to describe the problem-solving process.
Figure~\ref{A.3} presents the Xwin-style response with a more detailed calculation process.

\begin{figure}[H]
\begin{tcolorbox}[colback=blue!3!white, colframe=RoyalPurple!55!Aquamarine!100!, title=Example A.1: MetaMathQA-Style Query-Response Pair]

\textbf{Query:} If $x$ is a positive number such that \[\sqrt{8x}\cdot\sqrt{10x}\cdot\sqrt{3x}\cdot\sqrt{15x}=15,\]find all possible values for $x$.

\hdashrule[0.5ex]{\linewidth}{1pt}{3pt}

\textbf{Response:} Let's think step by step.

We start with the equation given:
$\sqrt{8x} \cdot \sqrt{10x} \cdot \sqrt{3x} \cdot \sqrt{15x} = 15$

First, simplify the left-hand side using the property of square roots:
$\sqrt{8x} \cdot \sqrt{10x} \cdot \sqrt{3x} \cdot \sqrt{15x} = \sqrt{(8x) \cdot (10x) \cdot (3x) \cdot (15x)}$

Next, multiply the expressions under the square root:
$(8x) \cdot (10x) \cdot (3x) \cdot (15x) = 8 \cdot 10 \cdot 3 \cdot 15 \cdot x^4$

We calculate the numerical product:
\[
8 \cdot 10 \cdot 3 \cdot 15 = 8 \cdot 10 \cdot 45 = 8 \cdot 450 = 3600
\]

Thus, the expression simplifies to:
$\sqrt{3600x^4} = 15$

Now, rewrite the expression inside the square root:
$\sqrt{3600x^4} = \sqrt{3600} \cdot \sqrt{x^4}$

Calculate the square roots:
$\sqrt{3600} = 60\text{ and } \sqrt{x^4} = x^2$

Thus, the equation becomes:
$60x^2 = 15$

Solve for \(x^2\):
$x^2 = \frac{15}{60} = \frac{1}{4}$

Take the positive square root of both sides (since \(x\) is positive):
$x = \sqrt{\frac{1}{4}} = \frac{1}{2}$

The answer is $\frac{1}{2}$.
\end{tcolorbox}
\vspace{-1em}
\caption{An example of data formatted in the MetaMathQA-style.}
\label{A.1}
\end{figure}

\vspace{2mm}

\begin{figure}[H]
\begin{tcolorbox}[colback=blue!3!white, colframe=RoyalPurple!55!Aquamarine!100!, title=Example A.2: Evol-Style Query-Response Pair]

\textbf{Query:} Suppose $x$ represents a positive quantity, and it satisfies the following mathematical relationship: 
\[\sqrt{8x} \times \sqrt{10x} \times \sqrt{3x} \times \sqrt{15x} = 15.\] Determine every viable value of $x$ that conforms to this equation.

\hdashrule[0.5ex]{\linewidth}{1pt}{3pt}

\textbf{Response:} We have the equation\[\sqrt{8x} \times \sqrt{10x} \times \sqrt{3x} \times \sqrt{15x} = 15.\]
We can simplify the left-hand side by multiplying the square roots together. Since the product of square roots is equal to the square root of the product of the values under the square roots (as long as the quantities under the square roots are positive, which they are since $x$ is positive), we have:$\sqrt{8x \cdot 10x \cdot 3x \cdot 15x} = 15.$
We can then multiply the numbers under the square root together:$\sqrt{8 \cdot 10 \cdot 3 \cdot 15 \cdot x^4} = 15.$
Next, we simplify the multiplication under the square root:$\sqrt{8 \cdot 10 \cdot 3 \cdot 15 \cdot x^4} = \sqrt{3600 \cdot x^4}.$
Notice that $3600 = 60^2$, so we can simplify the square root as follows:\[\sqrt{3600 \cdot x^4} = \sqrt{60^2 \cdot x^4}.\]
Using the property of the square root that $\sqrt{a^2} = a$ for any positive $a$, we get:\[\sqrt{60^2 \cdot x^4} = 60x^2.\]Now we can equate the simplified square root expression to the right-hand side of the original equation:$60x^2 = 15.$
To solve for $x^2$, we divide both sides of the equation by 60:
$x^2 = \frac{15}{60}.$
Simplify the fraction:
$x^2 = \frac{1}{4}.$
To solve for $x$, we take the square root of both sides, remembering that we're looking for the positive value of $x$ (since $x$ represents a positive quantity):$x = \sqrt{\frac{1}{4}}.$
The square root of $\frac{1}{4}$ is $\frac{1}{2}$, so we have:$x = \frac{1}{2}.$

The answer is $\frac{1}{2}.$
\end{tcolorbox}
\vspace{-1em}
\caption{An example of data formatted in the Evol-style.}
\label{A.2}
\end{figure}

\vspace{2mm}

\begin{figure}[H]
\begin{tcolorbox}[colback=blue!3!white, colframe=RoyalPurple!55!Aquamarine!100!, title=Example A.3: Xwin-Style Query-Response Pair]

\textbf{Query:} If \( x \) is a positive number such that 
\[
\sqrt{8x} \cdot \sqrt{10x} \cdot \sqrt{3x} \cdot \sqrt{15x} = 15,
\]
find all possible values for $x$.

\hdashrule[0.5ex]{\linewidth}{1pt}{3pt}

\textbf{Response:} Given the equation:
$\sqrt{8 x} \cdot \sqrt{10x} \cdot \sqrt{3 x} \cdot \sqrt{15 x}=15$

First, simplify the expression on the left side. We know that $\sqrt{a} \cdot \sqrt{b}=\sqrt{a b}$, so we can combine the square roots:
$\sqrt{8 x \cdot 10 x \cdot 3 x \cdot 15 x}$

Next, calculate the product inside the square root:
$8 x \cdot 10 x \cdot 3 x \cdot 15 x=8 \cdot 10 \cdot 3 \cdot 15 \cdot x^4$

Simplify the constants:
$8 \cdot 10=80$, $80 \cdot 3=240$, $240 \cdot 15=3600$.
So we have:
$8 x \cdot 10 x \cdot 3 x \cdot 15 x=3600 x^4$

Therefore, the equation becomes:
$\sqrt{3600 x^4}=24$
Since $\sqrt{3600 x^4}=\sqrt{3600} \cdot \sqrt{x^4}$, and $\sqrt{x^4}=x^2$, we get:
$\sqrt{3600} \cdot x^2=15$

Isolate $x^2$ by dividing both sides by $\sqrt{3600}$: $x^2=\frac{15}{\sqrt{3600}}$

Simplify $\sqrt{3600}$: $\sqrt{3600}=60$

So:
$x^2=\frac{15}{60}=\frac{1}{4}$

Finally, solve for $x$: $x=\sqrt{\frac{1}{4}}$

Upon checking if further simplification is possible, we find:
$x=\frac{1}{2}.$

The answer is $\frac{1}{2}.$

\end{tcolorbox}
\vspace{-1em}
\caption{An example of data formatted in the Xwin-style.}
\label{A.3}
\end{figure}

\section{Case Studies with Correct Answers Presented in Incorrect Formats} 
\label{sp cases of ans}
\begin{itemize}
    \item Different formats of the final answer but with the same value.
\begin{mdframed}[backgroundcolor=blue!3!white, linewidth=1.5pt, linecolor=black, roundcorner=10pt]
\noindent
\textbf{Ground Truth:} 0.24

\textbf{Response:} \dots The answer is 24\%
\end{mdframed}

\begin{mdframed}[backgroundcolor=blue!3!white, linewidth=1.5pt, linecolor=black, roundcorner=10pt]
\noindent
\textbf{Ground Truth:} $\sqrt{2}, \sqrt{3}$

\textbf{Response:} \dots The answer is $\sqrt{3}, \sqrt{2}$
\end{mdframed}

\begin{mdframed}[backgroundcolor=blue!3!white, linewidth=1.5pt, linecolor=black, roundcorner=10pt]
\noindent
\textbf{Ground Truth:} $\frac{2+\sqrt{2}}{4}$

\textbf{Response:} \dots The answer is $\frac{1}{2} + \frac{\sqrt{2}}{4}$
\end{mdframed}

\begin{mdframed}[backgroundcolor=blue!3!white, linewidth=1.5pt, linecolor=black, roundcorner=10pt]
\noindent
\textbf{Ground Truth:} $\backslash\backslash$text\{odd\}

\textbf{Response:} \dots The answer is $\backslash$"odd$\backslash$".
\end{mdframed}

\newpage
\item Unexpected format for presenting the final answer, such as rephrasing the prefix "\textbackslash nThe answer is " or including extra words before, in, or after "\textbackslash nThe answer is ".
\begin{mdframed}[backgroundcolor=blue!3!white, linewidth=1.5pt, linecolor=black, roundcorner=10pt]
\noindent
\textbf{Ground Truth:} 1, 2

\textbf{Response:} \dots The correct answer is 1, 2
\end{mdframed}

\begin{mdframed}[backgroundcolor=blue!3!white, linewidth=1.5pt, linecolor=black, roundcorner=10pt]
\noindent
\textbf{Ground Truth:} 19

\textbf{Response:} \dots The correct answer is 19, but this is based on an assumption that \dots
\end{mdframed}

\begin{mdframed}[backgroundcolor=blue!3!white, linewidth=1.5pt, linecolor=black, roundcorner=10pt]
\noindent
\textbf{Ground Truth:} 2

\textbf{Response:} \dots The value of x is 2
\end{mdframed}

\begin{mdframed}[backgroundcolor=blue!3!white, linewidth=1.5pt, linecolor=black, roundcorner=10pt]
\noindent
\textbf{Ground Truth:} 24.01

\textbf{Response:} \dots The answer is $x = \frac{2401}{100}= 24.01$
\end{mdframed}

\end{itemize}

\section{Performance Analysis in Stage 2 of the Data Synthesis pipeline} 
\label{model perf on hard lv}
Table~\ref{detailed level wise acc table} illustrates the relationship between data size in stage 2 of the data synthesis pipeline and the model performance. As we generate more hard synthetic problems in stage 2 of our data synthesis pipeline, the fine-tuned LLM models show gradual improvement in handling hard problems~(Level 3-5) on the MATH benchmark.

\begingroup
\setlength{\tabcolsep}{3pt} 
\renewcommand{\arraystretch}{1} 
\begin{table}[H]
\centering
\adjustbox{max width=\textwidth}{
\begin{tabular}{lcccccc}
\toprule
\multirow{2}{*}{\textbf{Base Model}} & \multirow{2}{*}{\textbf{{Dataset Size}}} & \multicolumn{5}{c}{\textbf{Difficulty Levels in MATH(\%)}} \\
\cmidrule(l){3-7}
& & \textbf{Level-1} & \textbf{Level-2} & \textbf{Level-3} & \textbf{Level-4} & \textbf{Level-5} \\
\midrule
LLaMA2-7B & 7.5K       & 17.85 & 8.39  & 4.77  & 3.05  & 0.91 \\
Mistral-7B & 7.5K      & 37.99 & 25.17 & 15.12 & 8.48  & 2.49 \\
DeepSeekMath-7B & 7.5K & 64.07 & 46.76 & 37.84 & 24.63 & 10.73 \\
\midrule
LLaMA2-7B & 1.0M      & 75.29 & 55.03 & 44.56 & 31.22 & 13.75 \\
Mistral-7B & 1.0M     & 80.55 & 63.31 & 53.05 & 38.47 & 19.18 \\
DeepSeekMath-7B & 1.0M & 79.18 & 62.30 & 54.82 & 40.44 & 19.71 \\
\midrule
LLaMA2-7B & 2.1M      & 78.03 & 60.29 & 48.19 & 35.09 & 19.56 \\
Mistral-7B & 2.1M     & 80.78 & 66.33 & 55.53 & 41.52 & 21.45 \\
DeepSeekMath-7B & 2.1M & 80.78 & 65.21 & 58.00 & 41.60 & 21.83 \\
\midrule
LLaMA2-7B & 2.1M + 0.1M (hard) & 78.03 & 62.19 & 48.89 & 36.66 & 17.98 \\
Mistral-7B & 2.1M + 0.1M (hard) & 81.01 & 67.45 & 58.44 & 45.22 & 21.53 \\
DeepSeekMath-7B & 2.1M + 0.1M (hard) & 84.90 & 67.45 & 57.91 & 44.07 & 21.22 \\
\midrule
LLaMA2-7B & 2.1M + 0.2M (hard) & 78.95 & 61.41 & 51.11 & 39.29 & 18.66 \\
Mistral-7B & 2.1M + 0.2M (hard) & 83.52 & 68.90 & 59.50 & 46.05 & 22.21 \\
DeepSeekMath-7B & 2.1M + 0.2M (hard) & 82.84 & 68.46 & 57.91 & 42.50 & 23.41 \\
\midrule
LLaMA2-7B & 2.1M + 0.4M (hard) & 78.03 & 62.42 & 52.87 & 37.48 & 18.73 \\
Mistral-7B & 2.1M + 0.4M (hard) & 83.52 & 67.56 & 60.65 & 44.89 & 25.08 \\
DeepSeekMath-7B & 2.1M + 0.4M (hard) & 82.84 & 67.23 & 58.71 & 42.01 & 21.30 \\
\midrule
LLaMA2-7B & 7.5k + 0.4M (hard) & 63.16 & 43.96 & 34.39 & 24.46 & 10.20 \\
Mistral-7B & 7.5k + 0.4M (hard) & 71.62 & 57.27 & 48.72 & 34.60 & 16.99 \\
DeepSeekMath-7B & 7.5k + 0.4M (hard) & 81.01 & 61.97 & 51.90 & 37.07 & 18.05 \\
\midrule
GPT-4-Turbo & - & 82.84 & 73.38 & 65.34 & 52.88 & 34.06 \\
\bottomrule
\end{tabular}
}
\caption{Difficulty level-wise performance of different base LLMs in Skywork-Math models and various sizes of SFT data on MATH. GPT-4-Turbo is evaluated using our designed grading criteria with 4-shot COT prompting.}
\label{detailed level wise acc table}
\end{table}
\endgroup

\section{Performance Analysis on MATH across Subjects} 
\label{MATH Problems Study}
Table~\ref{table type wise} presents the accuracy results on the MATH benchmark across various math subjects. The Skywork-Math models excel in the "Algebra" category as we scale up the synthetic SFT data. However, it struggles in some other math subjects, such as "Geometry", where the understanding of geometric concepts may be challenging for language LLM models.

\begin{table}[H]
\centering
\setlength{\tabcolsep}{3pt}
\scalebox{0.79}{
\begin{tabular}{lc|cccccccc}
\toprule
\textbf{Base Model} & \textbf{Dataset Size} & \textbf{Algebra} & \makecell{\textbf{Counting} \& \\ \textbf{Probability}} & \textbf{Geometry} & \makecell{\textbf{Intermediate} \\ \textbf{Algebra}} & \makecell{\textbf{Number} \\ \textbf{Theory}} & \textbf{Prealgebra} & \textbf{Precalculus} \\
\midrule
LLaMA2-7B & 7.5K & 6.66 & 4.01 & 3.34 & 1.33 & 3.89 & 11.71 & 1.10 \\
Mistral-7B & 7.5K & 21.65 & 9.07 & 9.19 & 3.77 & 8.89 & 28.01 & 3.85 \\
DeepSeekMath-7B & 7.5K & 52.15 & 21.10 & 19.42 & 11.07 & 26.67 & 50.63 & 9.71 \\
\midrule
LLaMA2-7B & 1.0M & 55.69 & 32.28 & 30.69 & 16.06 & 34.26 & 55.80 & 18.50 \\
Mistral-7B & 1.0M & 65.37 & 34.60 & 35.49 & 20.49 & 44.44 & 64.87 & 23.81 \\
DeepSeekMath-7B & 1.0M & 68.16 & 35.02 & 35.91 & 22.81 & 41.30 & 62.34 & 26.74 \\
\midrule
LLaMA2-7B & 2.1M & 62.09 & 36.50 & 33.82 & 18.38 & 41.85 & 59.93 & 21.79 \\
Mistral-7B & 2.1M & 66.72 & 40.93 & 38.00 & 23.48 & 43.70 & 68.08 & 26.19 \\
DeepSeekMath-7B & 2.1M & 69.92 & 41.14 & 36.33 & 25.03 & 45.19 & 65.56 & 29.30 \\
\midrule
LLaMA2-7B & 2.5M & 64.62 & 37.13 & 35.49 & 21.26 & 40.56 & 63.72 & 25.64 \\
Mistral-7B & 2.5M & 70.85 & 43.25 & 41.75 & 24.58 & 49.44 & 70.72 & 30.77 \\
DeepSeekMath-7B & 2.5M & 69.25 & 38.40 & 38.00 & 24.70 & 43.52 & 68.77 & 30.22\\
\bottomrule
\end{tabular}
}
\caption{MATH accuracies across subjects with different SFT data sizes.}
\label{table type wise}
\end{table}

\section{Effect of Model Maximum Length in Two Stages of the Data Synthesis Pipeline}
Table~\ref{table max length} presents the performance with three 7B base models in Skywork-Math model series with maximum lengths set of 512 and 2048 in the stage 1 \& 2 of the data synthesis pipeline.

\begingroup
\setlength{\tabcolsep}{3pt} 
\renewcommand{\arraystretch}{1} 
\begin{table*}[h!]
\centering
\adjustbox{max width=\textwidth}{
\begin{tabular}{lcccc}
\toprule
\textbf{Base Model} & \textbf{Data Synthesis Pipeline (Size)} & \textbf{Model Max Length} & \textbf{MATH(\%)} & \textbf{GSM8k(\%)} \\
\midrule\midrule
LLaMA2-7B & Stage 1 (2.1M) & 512 & 42.36 & 70.81 \\
LLaMA2-7B & Stage 1 (2.1M) & 2048 & 45.56 & 73.62 \\
\hline
Mistral-7B & Stage 1 (2.1M) & 512 & 47.14 & 81.05 \\
Mistral-7B & Stage 1 (2.1M) & 2048 & 49.1 & 83.25 \\
\hline
DeepSeekMath-7B & Stage 1 (2.1M) & 512 & 48.24 & 79.61 \\
DeepSeekMath-7B & Stage 1 (2.1M) & 2048 & 48.64 & 79.30 \\
\midrule\midrule
LLaMA2-7B & Stage 2 (2.5M) & 512  & 44.06 & 67.85 \\
LLaMA2-7B & Stage 2 (2.5M) & 2048 & 47.66 & 72.86 \\
\hline
Mistral-7B & Stage 2 (2.5M) & 512 & 50.56 & 82.41 \\
Mistral-7B & Stage 2 (2.5M) & 2048 & 51.22 & 83.93 \\
\hline
DeepSeekMath-7B & Stage 2 (2.5M) & 512 & 48.28 & 80.52 \\
DeepSeekMath-7B & Stage 2 (2.5M) & 2048 & 49.88 & 81.50 \\
\bottomrule
\end{tabular}
}
\caption{Model performance with different model maximum lengths.}
\label{table max length}
\end{table*}
\endgroup

\newpage
\section{More Experiments with Base LLM models after SFTing on the Skywork-Math Dataset}

As shown in Table~\ref{gen on diff model}, we conduct experiments with two additional pre-trained base LLM model. The results indicate that after SFTing on the Skywork-Math Dataset, both base models exhibit consistent performance improvement.

\input{tables/table-11}

%% file: tables/table-11.tex
\begingroup
\setlength{\tabcolsep}{3pt} 
\renewcommand{\arraystretch}{1} 
\begin{table}[h]
\centering
\adjustbox{max width=\textwidth}{
\begin{tabular}{lccc}
\toprule
\textbf{Base Model} & \textbf{Data Synthesis Pipeline~(Size)} & \textbf{GSM8K(\%)} & \textbf{MATH(\%)}\\
\midrule
LLaMA3-8B & - & 79.60 & 30.00 \\
LLaMA3-8B & Stage 1 (2.1M) & 80.82 & 50.34 \\
LLaMA3-8B & Stage 2 (2.5M) & 75.97 & 50.30 \\
\midrule
Llemma-7B & - & 36.40 & 18.00 \\
Llemma-7B & Stage 1 (2.1M) & 65.43 & 40.34 \\
Llemma-7B & Stage 2 (2.5M) & 66.03 & 40.08 \\
\bottomrule
\end{tabular}
}
\caption{Performance on LLaMA3-8B~\cite{llama3modelcard} and Llemma-7B~\cite{azerbayev2024llemma} base LLM models. We fine-tune the corresponding base LLM models using the Skywork-MathQA dataset in stage 1 and stage 2 of the data synthesis pipeline.}
\label{gen on diff model}
\end{table}
\endgroup

%% file: main.bbl
\begin{thebibliography}{72}
\providecommand{\natexlab}[1]{#1}
\providecommand{\url}[1]{\texttt{#1}}
\expandafter\ifx\csname urlstyle\endcsname\relax
  \providecommand{\doi}[1]{doi: #1}\else
  \providecommand{\doi}{doi: \begingroup \urlstyle{rm}\Url}\fi

\bibitem[Achiam et~al.(2023)Achiam, Adler, Agarwal, Ahmad, Akkaya, Aleman, Almeida, Altenschmidt, Altman, Anadkat, et~al.]{openai2024gpt4}
J.~Achiam, S.~Adler, S.~Agarwal, L.~Ahmad, I.~Akkaya, F.~L. Aleman, D.~Almeida, J.~Altenschmidt, S.~Altman, S.~Anadkat, et~al.
\newblock Gpt-4 technical report.
\newblock \emph{arXiv preprint arXiv:2303.08774}, 2023.

\bibitem[AI@Meta(2024)]{llama3modelcard}
AI@Meta.
\newblock Llama 3 model card.
\newblock 2024.
\newblock URL \url{https://github.com/meta-llama/llama3/blob/main/MODEL_CARD.md}.

\bibitem[Almazrouei et~al.(2023)Almazrouei, Alobeidli, Alshamsi, Cappelli, Cojocaru, Debbah, Goffinet, Hesslow, Launay, Malartic, et~al.]{almazrouei2023falcon}
E.~Almazrouei, H.~Alobeidli, A.~Alshamsi, A.~Cappelli, R.~Cojocaru, M.~Debbah, {\'E}.~Goffinet, D.~Hesslow, J.~Launay, Q.~Malartic, et~al.
\newblock The falcon series of open language models.
\newblock \emph{arXiv preprint arXiv:2311.16867}, 2023.

\bibitem[An et~al.(2023)An, Ma, Lin, Zheng, Lou, and Chen]{an2024learningLEMA}
S.~An, Z.~Ma, Z.~Lin, N.~Zheng, J.~Lou, and W.~Chen.
\newblock Learning from mistakes makes {LLM} better reasoner.
\newblock \emph{CoRR}, abs/2310.20689, 2023.
\newblock \doi{10.48550/ARXIV.2310.20689}.
\newblock URL \url{https://doi.org/10.48550/arXiv.2310.20689}.

\bibitem[Anil et~al.(2023)Anil, Dai, Firat, Johnson, Lepikhin, Passos, Shakeri, Taropa, Bailey, Chen, et~al.]{anil2023palm2}
R.~Anil, A.~M. Dai, O.~Firat, M.~Johnson, D.~Lepikhin, A.~Passos, S.~Shakeri, E.~Taropa, P.~Bailey, Z.~Chen, et~al.
\newblock Palm 2 technical report.
\newblock \emph{arXiv preprint arXiv:2305.10403}, 2023.

\bibitem[Anthropic(2024)]{claude3modelcard}
Anthropic.
\newblock The claude 3 model family: Opus, sonnet, haiku.
\newblock 2024.
\newblock URL \url{https://www-cdn.anthropic.com/de8ba9b01c9ab7cbabf5c33b80b7bbc618857627/Model_Card_Claude_3.pdf}.

\bibitem[Arora et~al.(2023)Arora, Singh, et~al.]{arora2023llmsJEEBench}
D.~Arora, H.~G. Singh, et~al.
\newblock Have llms advanced enough? a challenging problem solving benchmark for large language models.
\newblock \emph{arXiv preprint arXiv:2305.15074}, 2023.

\bibitem[Azerbayev et~al.(2023)Azerbayev, Schoelkopf, Paster, Santos, McAleer, Jiang, Deng, Biderman, and Welleck]{azerbayev2024llemma}
Z.~Azerbayev, H.~Schoelkopf, K.~Paster, M.~D. Santos, S.~McAleer, A.~Jiang, J.~Deng, S.~Biderman, and S.~Welleck.
\newblock Llemma: An open language model for mathematics.
\newblock In \emph{The 3rd Workshop on Mathematical Reasoning and AI at NeurIPS'23}, 2023.
\newblock URL \url{https://openreview.net/forum?id=0QHZrCWCH0}.

\bibitem[Bai et~al.(2022)Bai, Jones, Ndousse, Askell, Chen, DasSarma, Drain, Fort, Ganguli, Henighan, et~al.]{bai2022training}
Y.~Bai, A.~Jones, K.~Ndousse, A.~Askell, A.~Chen, N.~DasSarma, D.~Drain, S.~Fort, D.~Ganguli, T.~Henighan, et~al.
\newblock Training a helpful and harmless assistant with reinforcement learning from human feedback.
\newblock \emph{arXiv preprint arXiv:2204.05862}, 2022.

\bibitem[Bengio et~al.(2009)Bengio, Louradour, Collobert, and Weston]{bengio2009curriculum}
Y.~Bengio, J.~Louradour, R.~Collobert, and J.~Weston.
\newblock Curriculum learning.
\newblock In \emph{Proceedings of the 26th annual international conference on machine learning}, pages 41--48, 2009.

\bibitem[Brown et~al.(2020)Brown, Mann, Ryder, Subbiah, Kaplan, Dhariwal, Neelakantan, Shyam, Sastry, Askell, et~al.]{brown2020language}
T.~Brown, B.~Mann, N.~Ryder, M.~Subbiah, J.~D. Kaplan, P.~Dhariwal, A.~Neelakantan, P.~Shyam, G.~Sastry, A.~Askell, et~al.
\newblock Language models are few-shot learners.
\newblock \emph{Advances in neural information processing systems}, 33:\penalty0 1877--1901, 2020.

\bibitem[Cao et~al.(2023)Cao, Kang, Wang, and Sun]{cao2023instructionMining}
Y.~Cao, Y.~Kang, C.~Wang, and L.~Sun.
\newblock Instruction mining: When data mining meets large language model finetuning.
\newblock \emph{arXiv preprint arXiv}, 2307, 2023.

\bibitem[Casper et~al.(2023)Casper, Davies, Shi, Gilbert, Scheurer, Rando, Freedman, Korbak, Lindner, Freire, et~al.]{casper2023open}
S.~Casper, X.~Davies, C.~Shi, T.~K. Gilbert, J.~Scheurer, J.~Rando, R.~Freedman, T.~Korbak, D.~Lindner, P.~Freire, et~al.
\newblock Open problems and fundamental limitations of reinforcement learning from human feedback.
\newblock \emph{arXiv preprint arXiv:2307.15217}, 2023.

\bibitem[Chen et~al.(2022)Chen, Ma, Wang, and Cohen]{chen2023programofthoughts}
W.~Chen, X.~Ma, X.~Wang, and W.~W. Cohen.
\newblock Program of thoughts prompting: Disentangling computation from reasoning for numerical reasoning tasks.
\newblock \emph{CoRR}, abs/2211.12588, 2022.
\newblock URL \url{https://doi.org/10.48550/arXiv.2211.12588}.

\bibitem[Chiang et~al.(2023)Chiang, Li, Lin, Sheng, Wu, Zhang, Zheng, Zhuang, Zhuang, Gonzalez, et~al.]{vicuna2023}
W.-L. Chiang, Z.~Li, Z.~Lin, Y.~Sheng, Z.~Wu, H.~Zhang, L.~Zheng, S.~Zhuang, Y.~Zhuang, J.~E. Gonzalez, et~al.
\newblock Vicuna: An open-source chatbot impressing gpt-4 with 90\%* chatgpt quality, march 2023.
\newblock \emph{URL https://lmsys. org/blog/2023-03-30-vicuna}, 3\penalty0 (5), 2023.

\bibitem[Cobbe et~al.(2021)Cobbe, Kosaraju, Bavarian, Hilton, Nakano, Hesse, and Schulman]{cobbe2021trainingGSM8K}
K.~Cobbe, V.~Kosaraju, M.~Bavarian, J.~Hilton, R.~Nakano, C.~Hesse, and J.~Schulman.
\newblock Training verifiers to solve math word problems.
\newblock \emph{CoRR}, abs/2110.14168, 2021.
\newblock URL \url{https://arxiv.org/abs/2110.14168}.

\bibitem[DeepSeek-AI(2024)]{deepseekv2}
DeepSeek-AI.
\newblock Deepseek-v2: A strong, economical, and efficient mixture-of-experts language model, 2024.

\bibitem[Engstrom et~al.(2024)Engstrom, Feldmann, and Madry]{engstrom2024dsdm}
L.~Engstrom, A.~Feldmann, and A.~Madry.
\newblock Dsdm: Model-aware dataset selection with datamodels.
\newblock \emph{arXiv preprint arXiv:2401.12926}, 2024.

\bibitem[Gendron et~al.(2024)Gendron, Bao, Witbrock, and Dobbie]{gendron2024LLMnotReasoner}
G.~Gendron, Q.~Bao, M.~Witbrock, and G.~Dobbie.
\newblock Large language models are not strong abstract reasoners yet.
\newblock In \emph{ICLR 2024 Workshop: How Far Are We From AGI}, 2024.
\newblock URL \url{https://openreview.net/forum?id=Pc0fPGip78}.

\bibitem[Gou et~al.(2023)Gou, Shao, Gong, Yang, Huang, Duan, Chen, et~al.]{gou2023tora}
Z.~Gou, Z.~Shao, Y.~Gong, Y.~Yang, M.~Huang, N.~Duan, W.~Chen, et~al.
\newblock Tora: A tool-integrated reasoning agent for mathematical problem solving.
\newblock \emph{arXiv preprint arXiv:2309.17452}, 2023.

\bibitem[GPT-4o(2024)]{simpleevals}
GPT-4o.
\newblock Gpt-4o simple evals, 2024.
\newblock URL \url{https://github.com/openai/simple-evals}.

\bibitem[Gunasekar et~al.(2023)Gunasekar, Zhang, Aneja, Mendes, Del~Giorno, Gopi, Javaheripi, Kauffmann, de~Rosa, Saarikivi, et~al.]{gunasekar2023textbooks}
S.~Gunasekar, Y.~Zhang, J.~Aneja, C.~C.~T. Mendes, A.~Del~Giorno, S.~Gopi, M.~Javaheripi, P.~Kauffmann, G.~de~Rosa, O.~Saarikivi, et~al.
\newblock Textbooks are all you need.
\newblock \emph{arXiv preprint arXiv:2306.11644}, 2023.

\bibitem[Guo et~al.(2024)Guo, Zhu, Yang, Xie, Dong, Zhang, Chen, Bi, Wu, Li, et~al.]{guo2024deepseekcoder}
D.~Guo, Q.~Zhu, D.~Yang, Z.~Xie, K.~Dong, W.~Zhang, G.~Chen, X.~Bi, Y.~Wu, Y.~Li, et~al.
\newblock Deepseek-coder: When the large language model meets programming--the rise of code intelligence.
\newblock \emph{arXiv preprint arXiv:2401.14196}, 2024.

\bibitem[He et~al.(2024)He, Luo, Bai, Hu, Thai, Shen, Hu, Han, Huang, Zhang, et~al.]{he2024olympiadbench}
C.~He, R.~Luo, Y.~Bai, S.~Hu, Z.~L. Thai, J.~Shen, J.~Hu, X.~Han, Y.~Huang, Y.~Zhang, et~al.
\newblock Olympiadbench: A challenging benchmark for promoting agi with olympiad-level bilingual multimodal scientific problems.
\newblock \emph{arXiv preprint arXiv:2402.14008}, 2024.

\bibitem[Hendrycks et~al.(2021)Hendrycks, Burns, Kadavath, Arora, Basart, Tang, Song, and Steinhardt]{hendrycks2021measuring}
D.~Hendrycks, C.~Burns, S.~Kadavath, A.~Arora, S.~Basart, E.~Tang, D.~Song, and J.~Steinhardt.
\newblock Measuring mathematical problem solving with the {MATH} dataset.
\newblock In \emph{Thirty-fifth Conference on Neural Information Processing Systems Datasets and Benchmarks Track (Round 2)}, 2021.
\newblock URL \url{https://openreview.net/forum?id=7Bywt2mQsCe}.

\bibitem[Huang and Chang(2022)]{huang2022towards}
J.~Huang and K.~C.-C. Chang.
\newblock Towards reasoning in large language models: A survey.
\newblock \emph{arXiv preprint arXiv:2212.10403}, 2022.

\bibitem[Jiang et~al.(2023)Jiang, Sablayrolles, Mensch, Bamford, Chaplot, Casas, Bressand, Lengyel, Lample, Saulnier, et~al.]{jiang2023mistral}
A.~Q. Jiang, A.~Sablayrolles, A.~Mensch, C.~Bamford, D.~S. Chaplot, D.~d.~l. Casas, F.~Bressand, G.~Lengyel, G.~Lample, L.~Saulnier, et~al.
\newblock Mistral 7b.
\newblock \emph{arXiv preprint arXiv:2310.06825}, 2023.

\bibitem[Jiang et~al.(2024{\natexlab{a}})Jiang, Sablayrolles, Roux, Mensch, Savary, Bamford, Chaplot, Casas, Hanna, Bressand, et~al.]{jiang2024mixtral}
A.~Q. Jiang, A.~Sablayrolles, A.~Roux, A.~Mensch, B.~Savary, C.~Bamford, D.~S. Chaplot, D.~d.~l. Casas, E.~B. Hanna, F.~Bressand, et~al.
\newblock Mixtral of experts.
\newblock \emph{arXiv preprint arXiv:2401.04088}, 2024{\natexlab{a}}.

\bibitem[Jiang et~al.(2024{\natexlab{b}})Jiang, Shi, Yu, Liu, Zhang, Li, and Kwok]{jiang2024forwardbackwardFOBAR}
W.~Jiang, H.~Shi, L.~Yu, Z.~Liu, Y.~Zhang, Z.~Li, and J.~Kwok.
\newblock Forward-backward reasoning in large language models for mathematical verification, 2024{\natexlab{b}}.
\newblock URL \url{https://openreview.net/forum?id=GhYXocT75t}.

\bibitem[Kaplan et~al.(2020)Kaplan, McCandlish, Henighan, Brown, Chess, Child, Gray, Radford, Wu, and Amodei]{kaplan2020scalingLaws}
J.~Kaplan, S.~McCandlish, T.~Henighan, T.~B. Brown, B.~Chess, R.~Child, S.~Gray, A.~Radford, J.~Wu, and D.~Amodei.
\newblock Scaling laws for neural language models.
\newblock \emph{arXiv preprint arXiv:2001.08361}, 2020.

\bibitem[Kwon et~al.(2023)Kwon, Li, Zhuang, Sheng, Zheng, Yu, Gonzalez, Zhang, and Stoica]{kwon2023efficientvLLM}
W.~Kwon, Z.~Li, S.~Zhuang, Y.~Sheng, L.~Zheng, C.~H. Yu, J.~Gonzalez, H.~Zhang, and I.~Stoica.
\newblock Efficient memory management for large language model serving with pagedattention.
\newblock In \emph{Proceedings of the 29th Symposium on Operating Systems Principles}, pages 611--626, 2023.

\bibitem[Lan et~al.(2022)Lan, Wang, Zhang, Lan, Dai, Wang, Zhang, and Lim]{lan2022mwptoolkit}
Y.~Lan, L.~Wang, Q.~Zhang, Y.~Lan, B.~T. Dai, Y.~Wang, D.~Zhang, and E.-P. Lim.
\newblock Mwptoolkit: An open-source framework for deep learning-based math word problem solvers.
\newblock In \emph{Proceedings of the AAAI Conference on Artificial Intelligence}, volume~36, pages 13188--13190, 2022.

\bibitem[Lewkowycz et~al.(2022)Lewkowycz, Andreassen, Dohan, Dyer, Michalewski, Ramasesh, Slone, Anil, Schlag, Gutman-Solo, Wu, Neyshabur, Gur-Ari, and Misra]{lewkowycz2022minerva}
A.~Lewkowycz, A.~J. Andreassen, D.~Dohan, E.~Dyer, H.~Michalewski, V.~V. Ramasesh, A.~Slone, C.~Anil, I.~Schlag, T.~Gutman-Solo, Y.~Wu, B.~Neyshabur, G.~Gur-Ari, and V.~Misra.
\newblock Solving quantitative reasoning problems with language models.
\newblock In A.~H. Oh, A.~Agarwal, D.~Belgrave, and K.~Cho, editors, \emph{Advances in Neural Information Processing Systems}, 2022.
\newblock URL \url{https://openreview.net/forum?id=IFXTZERXdM7}.

\bibitem[Li et~al.(2024)Li, Wang, Hu, Wei, Zheng, Hu, Zhang, and Peng]{li2024common}
C.~Li, W.~Wang, J.~Hu, Y.~Wei, N.~Zheng, H.~Hu, Z.~Zhang, and H.~Peng.
\newblock Common 7b language models already possess strong math capabilities.
\newblock \emph{arXiv preprint arXiv:2403.04706}, 2024.

\bibitem[Li et~al.(2023)Li, Zhang, Li, Chen, Chen, Cheng, Wang, Zhou, and Xiao]{li2024quantitytoQuality}
M.~Li, Y.~Zhang, Z.~Li, J.~Chen, L.~Chen, N.~Cheng, J.~Wang, T.~Zhou, and J.~Xiao.
\newblock From quantity to quality: Boosting llm performance with self-guided data selection for instruction tuning.
\newblock \emph{CoRR}, abs/2308.12032, 2023.
\newblock URL \url{https://doi.org/10.48550/arXiv.2308.12032}.

\bibitem[Lu et~al.(2023)Lu, Peng, Cheng, Galley, Chang, Wu, Zhu, and Gao]{lu2023chameleon}
P.~Lu, B.~Peng, H.~Cheng, M.~Galley, K.-W. Chang, Y.~N. Wu, S.-C. Zhu, and J.~Gao.
\newblock Chameleon: Plug-and-play compositional reasoning with large language models.
\newblock In \emph{Thirty-seventh Conference on Neural Information Processing Systems}, 2023.
\newblock URL \url{https://openreview.net/forum?id=HtqnVSCj3q}.

\bibitem[Luo et~al.(2023)Luo, Sun, Xu, Zhao, Lou, Tao, Geng, Lin, Chen, and Zhang]{luo2023wizardmath}
H.~Luo, Q.~Sun, C.~Xu, P.~Zhao, J.~Lou, C.~Tao, X.~Geng, Q.~Lin, S.~Chen, and D.~Zhang.
\newblock Wizardmath: Empowering mathematical reasoning for large language models via reinforced evol-instruct.
\newblock \emph{CoRR}, abs/2308.09583, 2023.
\newblock URL \url{https://doi.org/10.48550/arXiv.2308.09583}.

\bibitem[Ni et~al.(2024)Ni, Gong, Gou, Shen, Yang, Duan, and Chen]{ni2024exploring}
X.~Ni, Y.~Gong, Z.~Gou, Y.~Shen, Y.~Yang, N.~Duan, and W.~Chen.
\newblock Exploring the mystery of influential data for mathematical reasoning.
\newblock \emph{arXiv preprint arXiv:2404.01067}, 2024.

\bibitem[Paster et~al.(2024)Paster, Santos, Azerbayev, and Ba]{paster2023openwebmath}
K.~Paster, M.~D. Santos, Z.~Azerbayev, and J.~Ba.
\newblock Openwebmath: An open dataset of high-quality mathematical web text.
\newblock In \emph{The Twelfth International Conference on Learning Representations}, 2024.
\newblock URL \url{https://openreview.net/forum?id=jKHmjlpViu}.

\bibitem[Peng et~al.(2023)Peng, Wu, Allard, Kilpatrick, and Heidel]{gpt3.5}
A.~Peng, M.~Wu, J.~Allard, L.~Kilpatrick, and S.~Heidel.
\newblock Gpt-3.5 turbo fine-tuning and api updates.
\newblock 2023.

\bibitem[Rafailov et~al.(2024)Rafailov, Sharma, Mitchell, Manning, Ermon, and Finn]{rafailov2024direct}
R.~Rafailov, A.~Sharma, E.~Mitchell, C.~D. Manning, S.~Ermon, and C.~Finn.
\newblock Direct preference optimization: Your language model is secretly a reward model.
\newblock \emph{Advances in Neural Information Processing Systems}, 36, 2024.

\bibitem[Saxton et~al.(2019)Saxton, Grefenstette, Hill, and Kohli]{saxton2019analysingDeepmindMath}
D.~Saxton, E.~Grefenstette, F.~Hill, and P.~Kohli.
\newblock Analysing mathematical reasoning abilities of neural models.
\newblock \emph{arXiv preprint arXiv:1904.01557}, 2019.

\bibitem[Scao et~al.(2022)Scao, Fan, Akiki, Pavlick, Ilic, Hesslow, Castagné, Luccioni, Yvon, Gallé, Tow, Rush, Biderman, Webson, Ammanamanchi, Wang, Sagot, Muennighoff, del Moral, Ruwase, Bawden, Bekman, McMillan-Major, Beltagy, Nguyen, Saulnier, Tan, Suarez, Sanh, Laurençon, Jernite, Launay, Mitchell, Raffel, Gokaslan, Simhi, Soroa, Aji, Alfassy, Rogers, Nitzav, Xu, Mou, Emezue, Klamm, Leong, van Strien, Adelani, and et~al.]{workshop2023bloom}
T.~L. Scao, A.~Fan, C.~Akiki, E.~Pavlick, S.~Ilic, D.~Hesslow, R.~Castagné, A.~S. Luccioni, F.~Yvon, M.~Gallé, J.~Tow, A.~M. Rush, S.~Biderman, A.~Webson, P.~S. Ammanamanchi, T.~Wang, B.~Sagot, N.~Muennighoff, A.~V. del Moral, O.~Ruwase, R.~Bawden, S.~Bekman, A.~McMillan-Major, I.~Beltagy, H.~Nguyen, L.~Saulnier, S.~Tan, P.~O. Suarez, V.~Sanh, H.~Laurençon, Y.~Jernite, J.~Launay, M.~Mitchell, C.~Raffel, A.~Gokaslan, A.~Simhi, A.~Soroa, A.~F. Aji, A.~Alfassy, A.~Rogers, A.~K. Nitzav, C.~Xu, C.~Mou, C.~Emezue, C.~Klamm, C.~Leong, D.~van Strien, D.~I. Adelani, and et~al.
\newblock Bloom: A 176b-parameter open-access multilingual language model.
\newblock \emph{CoRR}, abs/2211.05100, 2022.
\newblock URL \url{https://doi.org/10.48550/arXiv.2211.05100}.

\bibitem[Sener and Savarese(2017)]{sener2018activeCoreSet}
O.~Sener and S.~Savarese.
\newblock Active learning for convolutional neural networks: A core-set approach.
\newblock \emph{arXiv preprint arXiv:1708.00489}, 2017.

\bibitem[Shao et~al.(2024)Shao, Wang, Zhu, Xu, Song, Zhang, Li, Wu, and Guo]{shao2024deepseekmath}
Z.~Shao, P.~Wang, Q.~Zhu, R.~Xu, J.~Song, M.~Zhang, Y.~Li, Y.~Wu, and D.~Guo.
\newblock Deepseekmath: Pushing the limits of mathematical reasoning in open language models.
\newblock \emph{arXiv preprint arXiv:2402.03300}, 2024.

\bibitem[Shen et~al.(2023)Shen, Jin, Huang, Liu, Dong, Guo, Wu, Liu, and Xiong]{shen2023large}
T.~Shen, R.~Jin, Y.~Huang, C.~Liu, W.~Dong, Z.~Guo, X.~Wu, Y.~Liu, and D.~Xiong.
\newblock Large language model alignment: A survey.
\newblock \emph{arXiv preprint arXiv:2309.15025}, 2023.

\bibitem[Soviany et~al.(2022)Soviany, Ionescu, Rota, and Sebe]{soviany2022curriculum}
P.~Soviany, R.~T. Ionescu, P.~Rota, and N.~Sebe.
\newblock Curriculum learning: A survey.
\newblock \emph{International Journal of Computer Vision}, 130\penalty0 (6):\penalty0 1526--1565, 2022.

\bibitem[Sun et~al.(2023)Sun, Zheng, Xie, Liu, Chu, Qiu, Xu, Ding, Li, Geng, et~al.]{sun2023survey}
J.~Sun, C.~Zheng, E.~Xie, Z.~Liu, R.~Chu, J.~Qiu, J.~Xu, M.~Ding, H.~Li, M.~Geng, et~al.
\newblock A survey of reasoning with foundation models.
\newblock \emph{arXiv preprint arXiv:2312.11562}, 2023.

\bibitem[Taori et~al.(2023)Taori, Gulrajani, Zhang, Dubois, Li, Guestrin, Liang, and Hashimoto]{alpaca}
R.~Taori, I.~Gulrajani, T.~Zhang, Y.~Dubois, X.~Li, C.~Guestrin, P.~Liang, and T.~B. Hashimoto.
\newblock Stanford alpaca: An instruction-following llama model.
\newblock \url{https://github.com/tatsu-lab/stanford_alpaca}, 2023.

\bibitem[Toshniwal et~al.(2024)Toshniwal, Moshkov, Narenthiran, Gitman, Jia, and Gitman]{toshniwal2024openmathinstruct1}
S.~Toshniwal, I.~Moshkov, S.~Narenthiran, D.~Gitman, F.~Jia, and I.~Gitman.
\newblock Openmathinstruct-1: A 1.8 million math instruction tuning dataset.
\newblock \emph{arXiv preprint arXiv:2402.10176}, 2024.

\bibitem[Touvron et~al.(2023)Touvron, Martin, Stone, Albert, Almahairi, Babaei, Bashlykov, Batra, Bhargava, Bhosale, Bikel, Blecher, Canton-Ferrer, Chen, Cucurull, Esiobu, Fernandes, Fu, Fu, Fuller, Gao, Goswami, Goyal, Hartshorn, Hosseini, Hou, Inan, Kardas, Kerkez, Khabsa, Kloumann, Korenev, Koura, Lachaux, Lavril, Lee, Liskovich, Lu, Mao, Martinet, Mihaylov, Mishra, Molybog, Nie, Poulton, Reizenstein, Rungta, Saladi, Schelten, Silva, Smith, Subramanian, Tan, Tang, Taylor, Williams, Kuan, Xu, Yan, Zarov, Zhang, Fan, Kambadur, Narang, Rodriguez, Stojnic, Edunov, and Scialom]{touvron2023llama}
H.~Touvron, L.~Martin, K.~Stone, P.~Albert, A.~Almahairi, Y.~Babaei, N.~Bashlykov, S.~Batra, P.~Bhargava, S.~Bhosale, D.~Bikel, L.~Blecher, C.~Canton-Ferrer, M.~Chen, G.~Cucurull, D.~Esiobu, J.~Fernandes, J.~Fu, W.~Fu, B.~Fuller, C.~Gao, V.~Goswami, N.~Goyal, A.~Hartshorn, S.~Hosseini, R.~Hou, H.~Inan, M.~Kardas, V.~Kerkez, M.~Khabsa, I.~Kloumann, A.~Korenev, P.~S. Koura, M.-A. Lachaux, T.~Lavril, J.~Lee, D.~Liskovich, Y.~Lu, Y.~Mao, X.~Martinet, T.~Mihaylov, P.~Mishra, I.~Molybog, Y.~Nie, A.~Poulton, J.~Reizenstein, R.~Rungta, K.~Saladi, A.~Schelten, R.~Silva, E.~M. Smith, R.~Subramanian, X.~E. Tan, B.~Tang, R.~Taylor, A.~Williams, J.~X. Kuan, P.~Xu, Z.~Yan, I.~Zarov, Y.~Zhang, A.~Fan, M.~Kambadur, S.~Narang, A.~Rodriguez, R.~Stojnic, S.~Edunov, and T.~Scialom.
\newblock Llama 2: Open foundation and fine-tuned chat models.
\newblock \emph{CoRR}, abs/2307.09288, 2023.
\newblock URL \url{https://doi.org/10.48550/arXiv.2307.09288}.

\bibitem[Wang et~al.(2022)Wang, Wei, Schuurmans, Le, Chi, Narang, Chowdhery, and Zhou]{wang2023selfconsistency}
X.~Wang, J.~Wei, D.~Schuurmans, Q.~Le, E.~Chi, S.~Narang, A.~Chowdhery, and D.~Zhou.
\newblock Self-consistency improves chain of thought reasoning in language models.
\newblock \emph{arXiv preprint arXiv:2203.11171}, 2022.

\bibitem[Wang et~al.(2024)Wang, Hu, Lu, Zhu, Zhang, Subramaniam, Loomba, Zhang, Sun, and Wang]{wang2024scibench}
X.~Wang, Z.~Hu, P.~Lu, Y.~Zhu, J.~Zhang, S.~Subramaniam, A.~R. Loomba, S.~Zhang, Y.~Sun, and W.~Wang.
\newblock Scibench: Evaluating college-level scientific problem-solving abilities of large language models, 2024.
\newblock URL \url{https://openreview.net/forum?id=u6jbcaCHqO}.

\bibitem[Wei et~al.(2022{\natexlab{a}})Wei, Tay, Bommasani, Raffel, Zoph, Borgeaud, Yogatama, Bosma, Zhou, Metzler, et~al.]{wei2022emergent}
J.~Wei, Y.~Tay, R.~Bommasani, C.~Raffel, B.~Zoph, S.~Borgeaud, D.~Yogatama, M.~Bosma, D.~Zhou, D.~Metzler, et~al.
\newblock Emergent abilities of large language models.
\newblock \emph{arXiv preprint arXiv:2206.07682}, 2022{\natexlab{a}}.

\bibitem[Wei et~al.(2022{\natexlab{b}})Wei, Wang, Schuurmans, Bosma, Xia, Chi, Le, Zhou, et~al.]{wei2023chainofthought}
J.~Wei, X.~Wang, D.~Schuurmans, M.~Bosma, F.~Xia, E.~Chi, Q.~V. Le, D.~Zhou, et~al.
\newblock Chain-of-thought prompting elicits reasoning in large language models.
\newblock \emph{Advances in neural information processing systems}, 35:\penalty0 24824--24837, 2022{\natexlab{b}}.

\bibitem[Wei et~al.(2023{\natexlab{a}})Wei, Luan, Liu, Dong, and Wang]{wei2023cmath}
T.~Wei, J.~Luan, W.~Liu, S.~Dong, and B.~Wang.
\newblock Cmath: can your language model pass chinese elementary school math test?
\newblock \emph{arXiv preprint arXiv:2306.16636}, 2023{\natexlab{a}}.

\bibitem[Wei et~al.(2023{\natexlab{b}})Wei, Zhao, Zhang, Zhu, Wang, Yang, Li, Cheng, L{\"u}, Hu, et~al.]{wei2023skywork}
T.~Wei, L.~Zhao, L.~Zhang, B.~Zhu, L.~Wang, H.~Yang, B.~Li, C.~Cheng, W.~L{\"u}, R.~Hu, et~al.
\newblock Skywork: A more open bilingual foundation model.
\newblock \emph{arXiv preprint arXiv:2310.19341}, 2023{\natexlab{b}}.

\bibitem[Wei et~al.(2024)Wei, Zhu, Zhao, Cheng, Li, L{\"u}, Cheng, Zhang, Zhang, Zeng, et~al.]{wei2024skywork}
T.~Wei, B.~Zhu, L.~Zhao, C.~Cheng, B.~Li, W.~L{\"u}, P.~Cheng, J.~Zhang, X.~Zhang, L.~Zeng, et~al.
\newblock Skywork-moe: A deep dive into training techniques for mixture-of-experts language models.
\newblock \emph{arXiv preprint arXiv:2406.06563}, 2024.

\bibitem[Weng et~al.(2022)Weng, Zhu, Xia, Li, He, Liu, Sun, Liu, and Zhao]{weng2023largeSelfVerification}
Y.~Weng, M.~Zhu, F.~Xia, B.~Li, S.~He, S.~Liu, B.~Sun, K.~Liu, and J.~Zhao.
\newblock Large language models are better reasoners with self-verification.
\newblock \emph{arXiv preprint arXiv:2212.09561}, 2022.

\bibitem[Wu et~al.(2023)Wu, Qiu, Ross, Akyürek, Chen, Wang, Kim, Andreas, and Kim]{wu2024reasoningorreciting}
Z.~Wu, L.~Qiu, A.~Ross, E.~Akyürek, B.~Chen, B.~Wang, N.~Kim, J.~Andreas, and Y.~Kim.
\newblock Reasoning or reciting? exploring the capabilities and limitations of language models through counterfactual tasks.
\newblock \emph{CoRR}, abs/2307.02477, 2023.
\newblock URL \url{https://doi.org/10.48550/arXiv.2307.02477}.

\bibitem[Xu et~al.(2023)Xu, Sun, Zheng, Geng, Zhao, Feng, Tao, and Jiang]{xu2023wizardlm}
C.~Xu, Q.~Sun, K.~Zheng, X.~Geng, P.~Zhao, J.~Feng, C.~Tao, and D.~Jiang.
\newblock Wizardlm: Empowering large language models to follow complex instructions.
\newblock \emph{CoRR}, abs/2304.12244, 2023.
\newblock URL \url{https://doi.org/10.48550/arXiv.2304.12244}.

\bibitem[Xu et~al.(2024)Xu, Liu, Liu, Hou, Li, Zhang, Wang, Zeng, Du, Zhao, et~al.]{xu2024chatglmmath}
Y.~Xu, X.~Liu, X.~Liu, Z.~Hou, Y.~Li, X.~Zhang, Z.~Wang, A.~Zeng, Z.~Du, W.~Zhao, et~al.
\newblock Chatglm-math: Improving math problem-solving in large language models with a self-critique pipeline.
\newblock \emph{arXiv preprint arXiv:2404.02893}, 2024.

\bibitem[Xue et~al.(2024)Xue, Zheng, Fu, Ni, Zheng, Zhou, and You]{xue2024openmoe}
F.~Xue, Z.~Zheng, Y.~Fu, J.~Ni, Z.~Zheng, W.~Zhou, and Y.~You.
\newblock Openmoe: An early effort on open mixture-of-experts language models.
\newblock \emph{arXiv preprint arXiv:2402.01739}, 2024.

\bibitem[Yang et~al.(2023)Yang, Xiao, Wang, Zhang, Bian, Yin, Lv, Pan, Wang, Yan, et~al.]{yang2023baichuan2}
A.~Yang, B.~Xiao, B.~Wang, B.~Zhang, C.~Bian, C.~Yin, C.~Lv, D.~Pan, D.~Wang, D.~Yan, et~al.
\newblock Baichuan 2: Open large-scale language models.
\newblock \emph{arXiv preprint arXiv:2309.10305}, 2023.

\bibitem[Ying et~al.(2024)Ying, Zhang, Li, Zhou, Shao, Fei, Ma, Hong, Liu, Wang, et~al.]{ying2024internlm}
H.~Ying, S.~Zhang, L.~Li, Z.~Zhou, Y.~Shao, Z.~Fei, Y.~Ma, J.~Hong, K.~Liu, Z.~Wang, et~al.
\newblock Internlm-math: Open math large language models toward verifiable reasoning.
\newblock \emph{arXiv preprint arXiv:2402.06332}, 2024.

\bibitem[Yu et~al.(2024)Yu, Jiang, Shi, YU, Liu, Zhang, Kwok, Li, Weller, and Liu]{yu2023metamath}
L.~Yu, W.~Jiang, H.~Shi, J.~YU, Z.~Liu, Y.~Zhang, J.~Kwok, Z.~Li, A.~Weller, and W.~Liu.
\newblock Metamath: Bootstrap your own mathematical questions for large language models.
\newblock In \emph{The Twelfth International Conference on Learning Representations}, 2024.
\newblock URL \url{https://openreview.net/forum?id=N8N0hgNDRt}.

\bibitem[Yuan et~al.(2023)Yuan, Yuan, Tan, Wang, and Huang]{yuan2023largeArithmetic}
Z.~Yuan, H.~Yuan, C.~Tan, W.~Wang, and S.~Huang.
\newblock How well do large language models perform in arithmetic tasks?
\newblock \emph{CoRR}, abs/2304.02015, 2023.
\newblock URL \url{https://doi.org/10.48550/arXiv.2304.02015}.

\bibitem[Yue et~al.(2023)Yue, Qu, Zhang, Fu, Huang, Sun, Su, and Chen]{yue2023mammoth}
X.~Yue, X.~Qu, G.~Zhang, Y.~Fu, W.~Huang, H.~Sun, Y.~Su, and W.~Chen.
\newblock Mammoth: Building math generalist models through hybrid instruction tuning.
\newblock \emph{CoRR}, abs/2309.05653, 2023.
\newblock URL \url{https://doi.org/10.48550/arXiv.2309.05653}.

\bibitem[Zhang et~al.(2024)Zhang, Liu, Cherry, and Firat]{zhang2024scalingFinetuning}
B.~Zhang, Z.~Liu, C.~Cherry, and O.~Firat.
\newblock When scaling meets {LLM} finetuning: The effect of data, model and finetuning method.
\newblock In \emph{The Twelfth International Conference on Learning Representations}, 2024.
\newblock URL \url{https://openreview.net/forum?id=5HCnKDeTws}.

\bibitem[Zheng et~al.(2023)Zheng, Liu, Xie, Li, and Li]{zheng2023progressivehint}
C.~Zheng, Z.~Liu, E.~Xie, Z.~Li, and Y.~Li.
\newblock Progressive-hint prompting improves reasoning in large language models.
\newblock \emph{arXiv preprint arXiv:2304.09797}, 2023.

\bibitem[Zhong et~al.(2023)Zhong, Cui, Guo, Liang, Lu, Wang, Saied, Chen, and Duan]{zhong2023agieval}
W.~Zhong, R.~Cui, Y.~Guo, Y.~Liang, S.~Lu, Y.~Wang, A.~Saied, W.~Chen, and N.~Duan.
\newblock Agieval: A human-centric benchmark for evaluating foundation models.
\newblock \emph{arXiv preprint arXiv:2304.06364}, 2023.

\bibitem[Zhou et~al.(2023)Zhou, Liu, Xu, Iyer, Sun, Mao, Ma, Efrat, Yu, Yu, Zhang, Ghosh, Lewis, Zettlemoyer, and Levy]{zhou2023lima}
C.~Zhou, P.~Liu, P.~Xu, S.~Iyer, J.~Sun, Y.~Mao, X.~Ma, A.~Efrat, P.~Yu, L.~Yu, S.~Zhang, G.~Ghosh, M.~Lewis, L.~Zettlemoyer, and O.~Levy.
\newblock Lima: Less is more for alignment.
\newblock \emph{CoRR}, abs/2305.11206, 2023.
\newblock URL \url{https://doi.org/10.48550/arXiv.2305.11206}.

\end{thebibliography}
